\documentclass[STIX1COL]{ION-APA-Template}
\usepackage{tabularx}

\usepackage{threeparttable}

\usepackage{balance}
\makeatletter
\def\NAT@aysep{,}
\makeatother

\usepackage[%
backend=biber,
natbib=true,
style=apa,
]{biblatex}

\articletype{Preprint }%

\received{10-11-2022}
\revised{21-04-2023}
\accepted{23-06-2023 | Preprint version of the accepted submission to NAVIGATION, Journal of the Institute of Navigation }

\begin{document}

\title{Evaluation of the Benefits of Zero Velocity Update in Decentralized EKF-Based Cooperative Localization Algorithms for GNSS-Denied Multi-Robot Systems}

\author[1]{Cagri Kilic}

\author[2]{Eduardo Gutierrez}

\author[3]{Jason N. Gross}

\authormark{Kilic \textsc{et al}}

\address{\orgdiv{Department of Mechanical and Aerospace Engineering}, \orgname{West Virginia University}, \orgaddress{\state{WV}, \country{USA}}}



\corres{Cagri Kilic, 
\email{cagri.kilic@mail.wvu.edu}}

\abstract[Abstract]{
This paper proposes the cooperative use of zero velocity update (ZU) in a decentralized extended Kalman filter (DEKF) based localization algorithm for multi-robot systems. The filter utilizes inertial measurement unit (IMU), ultra-wideband (UWB), and odometry velocity measurements to improve the localization performance of the system in the presence of a GNSS-denied environment. The contribution of this work is to evaluate the benefits of using ZU in a DEKF-based localization algorithm. The algorithm is tested with real hardware in a video motion capture facility and a Robot Operating System (ROS) based simulation environment for unmanned ground vehicles (UGV). Both simulation and real-world experiments are performed to show the effectiveness of using ZU in one robot to reinstate the localization of other robots in a multi-robot system. Experimental results from GNSS-denied simulation and real-world environments show that using ZU with simple heuristics in the DEKF significantly improves the 3D localization accuracy. }



\maketitle


\section{Introduction}\label{sec1}

 Mobile robots rely on accurate localization estimates to perform certain tasks, such as exploration, navigation, object detection and tracking, map building, and autonomous movement through space. In a localization application, robots can enhance their ability to locate themselves accurately within the environment by fusing information from multiple sources. One common method of estimating positioning information is using the Global Navigation Satellite System (GNSS), which includes GPS and other similar systems. However, the availability of this system is often unreliable in urban, forested, and indoor areas because of obstructions that block signals from satellites~\citep{merry2019smartphone}.  

Given the challenges posed by limited GNSS availability in certain environments, cooperative localization emerges as a valuable alternative for mobile robots to achieve accurate positioning. Cooperation among multiple robots is desirable for many tasks, as the robots can perform several tasks more efficiently and robustly than a single robot~\citep{avinashmultirobot}. Cooperative localization is a technique in which multiple robots share information and perform relative measurements of one another to obtain a more accurate estimate of their location, compared to what would be possible by a single robot~\citep{gao2019cooperative}. Robots can achieve this level of cooperation when they detect each other and share state estimates and covariances in the presence of relative measurements, as shown by~\citet{luft2018recursive}. The individual state estimates of the robots are commonly obtained by fusing the measurements from proprioceptive (i.e., IMU) and exteroceptive (i.e., satellite signal receiver, laser scanner, camera) sensors.  

In a cooperative multi-robot system, an agent can still benefit from Global Navigation Satellite System (GNSS) information even without direct access to it if their counterparts have access to GNSS information~\citep{qu2011cooperative}. The fusion of IMU-based dead reckoning and visual odometry (VO) is useful in solving this problem in GNSS-denied/degraded environments and can be used cooperatively~\citep{pena2022VIOCoop}. However, relying on characteristics of the environment as in visual-based methods, accuracy is not reliable on uniform terrains with few landmarks.

ZU is widely used to aid pedestrian inertial navigation~\citep{foxlin2005, norrdine2016,kwakkel2008gnss,zhang2017adaptive} and is one of the important techniques of error suppression and compensation for high-precision positioning system~\citep{skog}. The main advantages of ZU for the localization task are that these updates can bound the velocity error, accurately calibrate the IMU sensor biases, and limit the rate of INS localization drift~\citep{grovebook}. ZU can be used in wheeled robots when stationary conditions are detected. ZU can be utilized passively, as an opportunistic navigational update, such as when wheeled robots need to stop for external reasons~\citep{edu2022pseudo}; or actively, with periodic stopping~\citep{kilic2019improved} or by deciding when to stop autonomously~\citep{kilic_slip-based_2021}. Since ZU can only be used when the robot is stationary, enforcing all the robots to stop in the system may be challenging in some cases. However, in a cooperative localization system, only some of the robots may need to stop actively, and the others can leverage ZU in an opportunistic way, such as when they need to stop for other reasons (e.g., avoiding obstacles, planning, waiting for pedestrians, stopping at traffic lights).  

In the case of multi-robot systems, cooperative localization can be classified into centralized and decentralized methods. In centralized methods, each robot in the system transmits its measurements to a central server, and the localization estimations of all robots are estimated on this server. This usually results in high communication and computational costs at the server~\citep{bailey2011decentralised}. Also, a failure of the central processing unit in a centralized method usually leads to catastrophic results. In addition, decentralized methods are more resilient to failures and  decrease the computational and communication costs~\citep{bailey2011decentralised}. For example, a multi-robot system may still function in the event of a malfunction of one individual robot in a decentralized method.

In decentralized localization methods, the impact of individual updates could be noticeably beneficial to the localization performance of the entire system. For example, having part of the multi-robot system able to perform GNSS updates could enhance the state estimates and the entire group. For example, suppose that some of the agents in the multi-robot system are in an area where the GNSS signal is interrupted, and others can obtain sufficient signals. In this situation, robots with adequate positioning information can share this information among the system during relative updates. This notion could easily be extrapolated to the use of other updates. For example, when a robot is in a stationary condition, it can perform ZU to calibrate the IMU biases to keep the INS-based localization reliable when other sensor measurements are not available (e.g., GNSS and VO), then the other robots in the system can benefit from this update even they do not use ZU. 

\section{Problem Statement}\label{sec2}
This paper explores the potential of leveraging the benefits of ZUs in decentralized cooperative localization. In this paper, we assume that each agent in a multi-robot system is able to perform INS-based dead-reckoning. The localization algorithm can leverage ZU in certain conditions depending on the robot type, such as landing~\citep{grovebook} and hovering~\citep{gross2019field} for aerial robots or stopping and using non-holonomicity~\citep{kilic2019improved} for ground robots. These specific conditions will allow the algorithm to perform ZU. We adopt a decentralized architecture for the filter estimator, which enables robots to decouple their states, reducing computational costs through distributed computation~\citep{bailey2011decentralised}. In order to perform relative ranging measurement updates, the robots can utilize Ultra Wide Band (UWB) sensors. This is done by coupling the states and covariances of the robots participating in the update. The individual robots take advantage of ZUs, which improve their localization performance in feature-poor areas without significant changes to robot operations. This enhancement, similar to the effect of using a GNSS update for a single robot, benefits the overall localization performance of the multi-robot system during relative updates.

It is assumed that all sensors have noise, and the robot's knowledge of its state is based on assumptions about the consistency of its representations of the real world. As a result, the error in the robot state can accumulate over time~\citep{thrun2005probabilistic,engelson1992error}. A large error accumulation increases the robot's uncertainty in localization estimation and can also generate a false belief of where it is located~\citep{choset2005principles}. In literature, the kidnapped robot problem is a case when a robot is moved to another position without being told~\citep{fox2001particle,engelson1992error}. Many authors have explored the kidnapped robot problem in many different contexts~\citep{yu2020deep,yi2011detection,bukhori2017detection,thrun2005probabilistic,choset2005principles}. In our case, similar to the kidnapped robot problem, we adopted lost robot cases, where the robots have significant position error and covariances without being moved to another position. Some of the lost robot cases can be observed when the primary localization sensors of the robot are disabled, or the error accumulation in the localization is larger than a sustainable level. In this study, we will focus on one specific implementation of this problem: keeping a robot's localization reliable to reinstate the localization of the other lost robots in a cooperative DEKF architecture.   

In our previous work, several schemes for deciding when to utilize the information from pseudo-measurements (e.g., ZU) to improve the localization performance in a DEKF based cooperative localization system is analyzed and compared in a simulation environment with ground robots \citep{edu2022pseudo}.

In this work, our specific contributions are summarized as follows:
\begin{itemize}
    \item We further generalize the method and significantly expand upon by implementing the algorithm into real-world experiments with small ground robots as presented in~\citet{edu2022pseudo}
    \item We add an additional algorithm to use velocity information from odometry (wheel and/or visual) in DEKF.
    \item We demonstrate the benefits of using ZU for the lost robot cases in a multi-robot DEKF system by qualitatively comparing the localization performance with video motion capture system solution. 
    \item We make our software and datasets publicly available\footnote{\url{https://github.com/wvu-navLab/coop_smart} }.
\end{itemize}

The rest of this paper is organized as follows. Section~\ref{Methodology} details and explains the implementation of the algorithm used. Section ~\ref{Experiments} describes the simulation and real-world experiments set up with the robots used. Section~\ref{Evaluation} provides insights into the results of simulation experiments. Finally, Section~\ref{Conclusion} provides contributions and insights for future works to improve the system.

\section{Methodology}\label{Methodology}
In this work, the base D-EKF algorithm is implemented closely following the work in~\citet{luft2016recursive}. This algorithm is embedded into ROS for localizing $N$ vehicles navigating in a GNSS-denied/degraded environment. Apart from the re-implementation of the base D-EKF algorithm into ROS, an error-state EKF is used instead of a total-state EKF. The reason to use an error-state EKF is to keep the accumulated error more sustainable over longer sequential predictions without measurement updates. Also, modeling the error states is more straightforward than the highly non-linear total states since the error states behave less complicated than the total state~\citep{roumeliotis1999circumventing, kilic2021planetary}. 

Additionally, the available sensor modalities are reduced such that the sensor fusion is done only using IMU, UWB, wheel encoders, and zero velocity updates. For instance, we assume that the robots cannot acquire bearing measurements of each other through exteroceptive sensors, are unable to identify any landmarks within the environment, and operate in a GNSS-denied setting. This differs from the approach presented in the work of~\citet{luft2016recursive}. The algorithm uses the IMU as the primary source for localization estimation. Since the INS-based localization is prone to drift over time, the estimation needs to be improved or corrected. Using wheel encoders and cameras for velocity information is widely utilized to improve the INS-based localization by sensor fusion techniques~\citep{cho2011dr,cobos2011vo}. GNSS updates can provide reliable information to correct for the localization drift; however, these updates are not always available. ZU can improve the localization estimation by calibrating the IMU and bounding the velocity error estimation, which can be available whenever the robot is stationary. The relative update is performed only using UWB-ranging information. This update is also utilized as a communication bridge between robots to share the state information pairwise.

In the error-state D-EKF architecture, each robot can perform three different algorithms to estimate and improve the state: 
\begin{enumerate}
    \item INS-based dead-reckoning, where each robot propagates its error state and updates its total state using the IMU measurements
    \item Private updates\footnote{Since the base D-EKF algorithm is implemented from the work in~\citet{luft2016recursive}, we opted to use the same naming convention with the cited work}, where each robot receives a measurement update, based on the sensor availability and robot-environment constraints, which is not shared directly with the group, (e.g., odometry velocity, ZU, and GNSS)

    \item Relative update$^2$, which is performed when two robots are within specified proximity, allowing for coupling the states, covariances, and cross-correlated values.
\end{enumerate}

These three algorithms are detailed in the following subsections. Fig.~\ref{architecture} shows the algorithms and the sensors used in the ROS system. 

 \begin{figure}[htb!] \label{architecture}
        \centering
         \includegraphics[width=\textwidth]{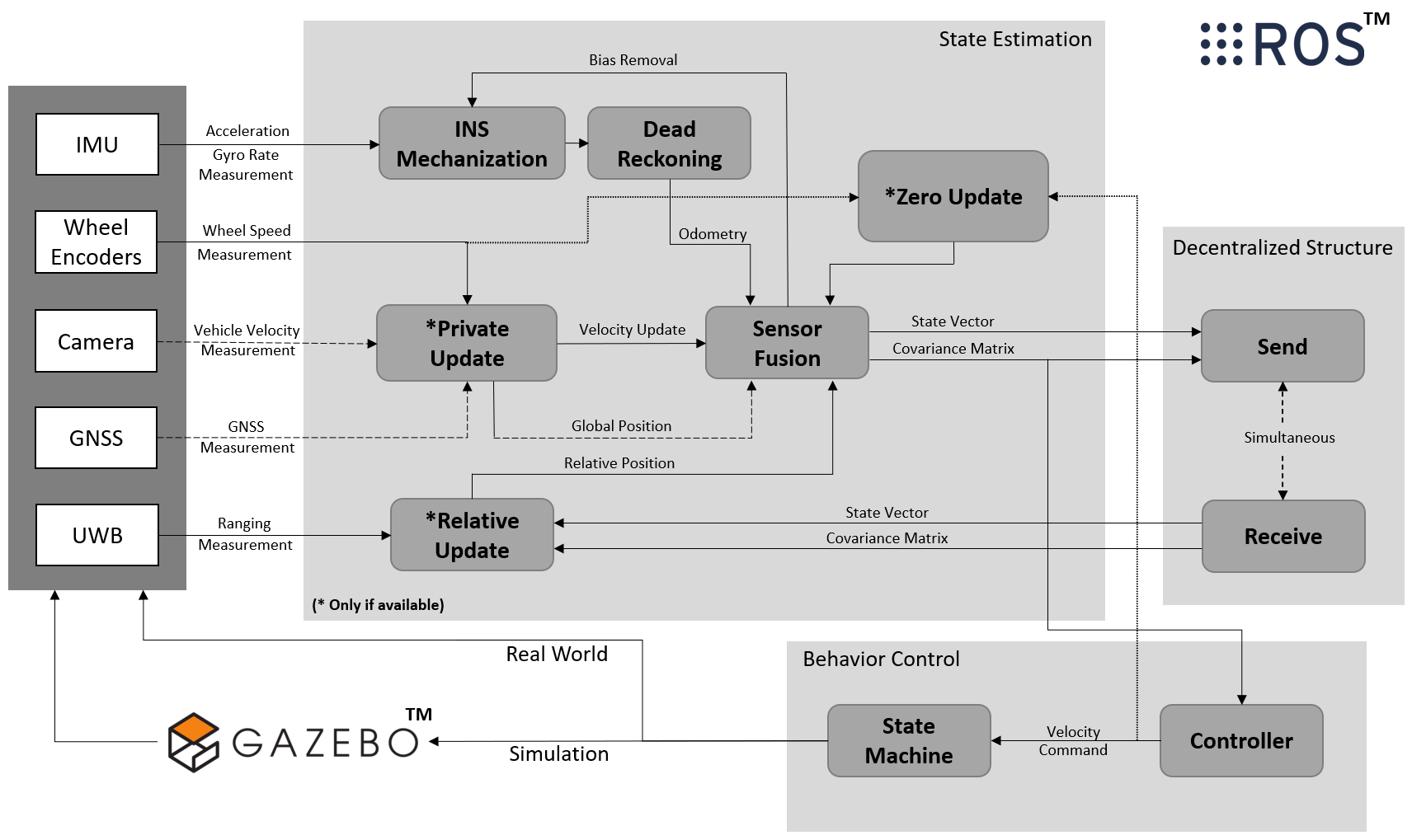}
        \caption{The architecture of the system. IMU measurements are used for the state propagation. Wheel encoder measurements are used to update the velocity and to trigger ZU. Camera-based visual odometry velocity update can optionally be used in the system to provide better estimation in feature rich environments. GNSS measurement update can be used to update the position and velocity (loosely coupled) if it is available. UWB measurement is used along with the state and covariance received from other robots to perform relative updates. In the state estimation block, dashed lines show the optional measurements and the dotted lines show the triggers for ZU. }  
        \label{architecture}
    \end{figure}

\subsection{INS-based Dead-Reckoning}
The INS implementation is the state propagation step of the error-state D-EKF implementation. The error-state formulation is in a local navigation frame (i.e., East-North-Up frame) based on the formulation in~\citet{grovebook} and~\citet{kilic_slip-based_2021}. This formulation is further simplified with the assumptions of having low velocity profiles of the robots which allows for neglecting the Earth's rotation and craft-rate terms in the INS mechanization. This is done to simplify INS initialization to not be dependent on the robot's absolute position on Earth and has limited performance impact with low-cost Micro-Electro-Mechanical Systems (MEMS) IMU-sensors.

The error-state vector, $\mathbf{x}_{err} \in \mathbb{R}^{15}$,  is formed in a local navigation frame, 
\begin{equation}
\label{errorstate}
\mathbf{x}_{err}^{n}={\biggl(
\delta\mathbf{ \Psi}_{nb}^{n} \ \ 
\delta\mathbf{v}_{eb}^{n} \ \ 
\delta\mathbf{r}_{b} \ \ 
\mathbf{b}_a \ \ 
\mathbf{b}_g
\biggr )}^{\mathbf{T}}
\end{equation}
where, $\delta\mathbf{ \Psi}_{nb}^{n}$ is the attitude error, $\delta\mathbf{ v}_{eb}^{n}$ is the velocity error, $\delta\mathbf{p}_{b}$ is the position error, $\mathbf{b}_a$ is the IMU acceleration bias, and $\mathbf{b}_g$ is the IMU gyroscope bias. Once the inertial biases $\mathbf{b}_a$ and $\mathbf{b}_g$ are estimated, they are used in our implementation by removing them from the raw IMU measurements before being used in the filter time propagation update. The error-state vector is assumed to be defined by~\eqref{errorstate} and the total state vector, $\mathbf{x} \in \mathbb{R}^{9}$, is 
\begin{equation}
\mathbf{x}^{n}={\biggl(
\mathbf{\Psi}_{nb}^{n} \ \ 
\mathbf{v}_{eb}^{n} \ \ 
\mathbf{r}_{b} 
\biggr )}^{\mathbf{T}} 
\end{equation}
where each of the nine total states correspond to the first nine error-states. In the inertial navigation equations, following the notation in \citep{grovebook}, the symbols $(-)$ and $(+)$ are employed to indicate the values at the beginning and end of the navigation equations processing cycle, respectively. The attitude update is given with the assumption of neglecting the rotation of Earth and transport rate as
\begin{equation}
\label{eq:su2}
\begin{aligned}
{\mathbf{C}_{b}^{n}}{(+)}\approx{} & {\mathbf{{C}}_{b}^{n}}{(-)}  \bigl (\mathbf{I}_3 + \mathbf{\Omega}_{ib}^{b} \Delta t_i \bigr )
\end{aligned}
\end{equation}
where  $\mathbf{C}_{b}^{n}$ is the coordinate transformation matrix from the body frame to the locally level frame, $\mathbf{I_3}$ is a 3-by-3 identity matrix, $\mathbf{\Omega}_{ib}^{b}$ is the skew symmetric matrix of the IMU angular rate measurement, and $\Delta t_i$ is the IMU sampling interval. The velocity update is given as,
\begin{equation}
\begin{aligned}
{\mathbf{v}_{eb}^{n}}{(+)} \approx {} & {\mathbf{v}_{eb}^{n}}{(-)} +(\mathbf{C}_{b}^{n}
\mathbf{a}_{IMU}+\mathbf{g}_{b}^{n} )\Delta t_i 
\end{aligned}
\end{equation}
where $\mathbf{v}_{eb}^{n}$ is the velocity update, $\mathbf{a}_{IMU}$ is the acceleration measurements from the IMU sensor, $\mathbf{g}_{b}^{n}$ is the gravity vector defined as $\mathbf{g}_{b}^{n} = {[
0, \ \ 
0, \ \ 
-9.81 
]}^{\mathbf{T}} $ .

The position update is given as
\begin{equation}
\mathbf{r}_b{(+)} \approx{} \mathbf{r}_b{(-)} -\frac{\Delta t_i}{2} \biggl({\mathbf{v}_{eb}^{n}}{(-)}  +{\mathbf{v}_{eb}^{n}}{(+)}  \biggr).
\end{equation}

The error-state dynamics is linearized and given as
\begin{equation}
\mathbf{F}=\left[\begin{array}{ccccc}
\mathbf{0}_{3}  & \mathbf{0}_{3} & \mathbf{0}_{3} & \mathbf{0}_{3} & \mathbf{C}_{b}^{n} \\
\wedge\left(-\mathbf{C}_{b}^{n}
\mathbf{a}_{IMU}\right)& \mathbf{0}_{3} & \mathbf{0}_{3}  & \mathbf{C}_{b}^{n} & \mathbf{0}_{3} \\
\mathbf{0}_{3} &\mathbf{I}_{3}  & \mathbf{0}_{3}  & \mathbf{0}_{3} & \mathbf{0}_{3} \\
\mathbf{0}_{3} & \mathbf{0}_{3} & \mathbf{0}_{3} & \mathbf{0}_{3} & \mathbf{0}_{3} \\
\mathbf{0}_{3} & \mathbf{0}_{3} & \mathbf{0}_{3} & \mathbf{0}_{3} & \mathbf{0}_{3}
\end{array}\right]
\end{equation}
where $\wedge$ is the skew-symmetric matrix of a vector. 

Then, the first order error-state transition matrix, $\Phi$, is given as 
\begin{equation}
\mathbf{\Phi} = \mathbf{I} + \mathbf{F} \Delta t    
\end{equation}
where it is used for propagating the error state vector, $\mathbf{x}_{err}$, the covariance, $\mathbf{P}$, and the cross-correlated terms, ${\mathbf{\sigma}}_{AB}$, between robots. The state estimates and covariance that are propagated over time are denoted using the superscript "$^-$", and those after the measurement updates later described in the following subsections are indicated using the superscript "$^+$". 
\begin{equation}
\mathbf{x}_{err}^{-}=\mathbf{\Phi} \mathbf{x}_{err}^{+}
\end{equation}
\begin{equation}
{\mathbf{P}}^{-}=\mathbf{\Phi} {\mathbf{P}}^{+} \mathbf{\Phi^{T}} + \mathbf{Q_{INS}}
\end{equation}
\begin{equation}
    \mathbf{\sigma}_{AB}^{-}= \mathbf{\Phi} \mathbf{\sigma}_{AB}^{+} 
\end{equation}
where $\mathbf{Q_{INS}}$ is the INS system noise covariance matrix which is generated by assuming the propagation intervals are small based on the work in~\citet{grovebook}. The input power spectral density for the error states is modeled by considering not only white noise in the accelerations and angular rates but also biases and scale factor errors inherent in inertial sensors. The construction of the $\mathbf{Q_{INS}}$ matrix involves accounting for bias and noise sources such as gyro in-run bias stability, angular random walk (ARW), accelerometer in-run bias stability, and velocity random walk (VRW). These are converted to their respective power spectral densities (PSDs) and scaled according to the time interval. The resulting covariance matrix, $\mathbf{Q_{INS}}$, is formed by populating the diagonal blocks with the computed PSDs for each noise source, ensuring an accurate representation of noise characteristics in the system.

\subsection{Private Updates}
Following the nomenclature in~\citet{luft2016recursive}, private updates are performed individually by each robot and are not shared with the group directly. These updates depend on the robot's individual relations with the environment. The information from some of these updates may not be available based on the mission environment profile. 
\subsubsection{GNSS and Motion Capture Update}
One common way to use the private updates is leveraging the GNSS update which are assumed to correct the localization drift and provide a reliable position estimate. As in our previous work~\citep{edu2022pseudo}, the GNSS update can be performed in a loosely coupled manner with the following structure. In this work, GNSS update is not used due to the indoor testing setting. However, we provided the structure of this update in the following for the sake of completeness.

The measurement innovation is given as 
\begin{equation} \label{measurement_gps}
    \mathbf{z}_{PU} = [ \mathbf{v}_{PU}-\mathbf{v}_{eb}^{n}, \mathbf{r}_{PU}-\mathbf{r}_{b}]
\end{equation}
where $\mathbf{v}_{PU}$ represents the velocity and $\mathbf{r}_{PU}$ represents the position measurement obtained from GNSS measurement. In our case, the velocity and position measurement coordinate frames are set to match with the corresponding state and the lever arm for the GNSS antenna is assumed to mount to the location of the IMU sensor. The Kalman gain for the private update is calculated as
\begin{equation}\label{kalman_gps}
    \mathbf{K}_{PU}={\mathbf{P}}^{-}\mathbf{H}_{PU}^{T}(\mathbf{H}_{PU} {\mathbf{P}}^{-} \mathbf{H}_{PU}^{T} +\mathbf{R}_{PU})^{-1}
\end{equation}
where $\textbf{H}_{PU}$ represents the Jacobian of the GNSS measurement model and can be given as
\begin{equation}
\mathbf{H}_{PU} =
\begin{bmatrix}
\mathbf{0}_{6x3} & \mathbf{-I}_{6x6} & \mathbf{0}_{6x6} 
\end{bmatrix}
\end{equation}
where $\textbf{I}$ represents the identity matrix. The measurement noise covariance matrix, $\textbf{R}_{PU}$, given as
\begin{equation}
\mathbf{R}_{PU} =  \mathrm{diag}(\sigma_{v_x}^2, \sigma_{v_y}^2, \sigma_{v_z}^2, \sigma_{r_x}^2, \sigma_{r_y}^2, \sigma_{r_z}^2)
\end{equation}
where  $\sigma_{v_x}^2$, $\sigma_{v_y}^2$, and $\sigma_{v_z}^2$ are the variances of the velocity measurement noise; $\sigma_{r_x}^2$, $\sigma_{r_y}^2$, and $\sigma_{r_z}^2$ are the variances of the position measurement noise in the $x$, $y$, and $z$ directions, respectively. Using the calculated Kalman gain, the error-state and covariance are updated as
\begin{equation}\label{state_gps}
\mathbf{x}_{err}^{+}=\mathbf{x}_{err}^{-}+\mathbf{K}_{PU}(\mathbf{z}_{PU}-\mathbf{H}_{PU}\mathbf{x}_{err}^{-})    
\end{equation}
\begin{equation}\label{covariance_gps}
{\mathbf{P}}^{+} =(\textbf{I}-\mathbf{K}_{PU}\mathbf{H}_{PU}){\mathbf{P}}^{-} (\textbf{I}-\mathbf{K}_{PU}\mathbf{H}_{PU})^{T} +\mathbf{K}_{PU} \mathbf{R}_{PU} \mathbf{K}_{PU}^{T} 
\end{equation}
Lastly, the cross-correlated terms with the rest of the robots in the system are updated as
\begin{equation}\label{correlation_gps}
    \mathbf{\sigma}_{AB}^{+} = (\textbf{I}-\mathbf{K}_{PU}\mathbf{H}_{PU})\mathbf{\sigma}_{AB}^{-}  (\textbf{I}-\mathbf{K}_{PU}\mathbf{H}_{PU})^{T} +\mathbf{K}_{PU} \mathbf{R}_{PU} \mathbf{K}_{PU}^{T} 
\end{equation}

where the $\mathbf{\sigma}_{AB}$ represents the cross-correlation terms from the individual robot to the rest of the robots in the system.

A video motion capture system (VICON) positioning update can be used as a private update to have reliable localization estimation for indoor cases and as a testing proxy for GNSS inside a laboratory setting. The VICON update can be used similar to the GNSS update framework. In our tests, we only used the VICON system to have a ground truth and also to initialize the robot position. Assuming the VICON solution frame matches the robot navigation frame,
\begin{equation} \label{measurement_vicon}
    \mathbf{z}_{vicon} = [\mathbf{r}_{vicon}-\mathbf{r}_{b}]
\end{equation}
where $\mathbf{r}_{vicon}$ represents the position measurement obtained from VICON measurement. The Jacobian of the measurement model and the measurement noise covariance matrix for VICON can be given as;
\begin{equation}
\mathbf{H}_{vicon}=\begin{bmatrix}
\mathbf{0}_{3x3} & \mathbf{0}_{3x3} & \mathbf{-I}_{3x3} & \mathbf{0}_{3x3} & \mathbf{0}_{3x3}  
\end{bmatrix}
\end{equation}
\begin{equation}
\mathbf{R}_{vicon} =\mathrm{diag} (\sigma_{vicon_x}^2,\sigma_{vicon_y}^2,\sigma_{vicon_z}^2)
\end{equation}
where $\sigma_{vicon_x}^2$, $\sigma_{vicon_y}^2$, and $\sigma_{vicon_z}^2$ are the variances of the VICON measurement noise.

\subsubsection{Odometry Velocity Update}
The odometry velocity update can be utilized to further improve the state estimation of individual robots. In this update, the algorithm only takes the velocity information from the associated sensor. For example, wheeled robots can use wheel encoders to obtain this information. Also, any external odometry solution can be utilized in this update such as the velocity information from the visual odometry or this update can be omitted if there is no sensor available.     

After the frame rotation between the velocity sensor to the navigation frame, i.e., $\mathbf{v}_{VU}$ is in the navigation frame, the measurement innovation is given as;
\begin{equation} \label{measurement_wo}
    \mathbf{z}_{VU} = [\mathbf{v}_{VU}-\mathbf{v}_{eb}^{n}]
\end{equation}
where $\mathbf{v}_{VU}$ represents the velocity measurement obtained from the odometry source. Updating the error-state, covariance, and cross-correlated terms follow the same structure through Equations~\ref{kalman_gps}-\ref{correlation_gps}. Jacobian of the measurement model, $\mathbf{H}_{VU}$, can be given as;
\begin{equation}
\mathbf{H}_{VU}=\begin{bmatrix}
\mathbf{0}_{3x3} & \mathbf{-I}_{3x3} & \mathbf{0}_{3x3} & \mathbf{0}_{3x3} & \mathbf{0}_{3x3}  
\end{bmatrix}
\end{equation}
and the measurement noise covariance matrix, $\mathbf{R}_{VU}$, can be constructed by the variances of the velocity measurement noise. 
\subsubsection{Zero Velocity Update}
In this work, ZU is applied using a combination of linear and angular velocity. To properly use this update, stationary conditions must be detected accurately. Otherwise, the rover's state yields incorrect updates leading to poor localization performance~{\citep{ramo}}. To detect stationary conditions, we used two different indicators, the velocity command provided by the autonomous controller that determines the movement of the robot and the wheel encoders measurements. We assume that robots do not slip under these conditions to determine stationary conditions and the robots do not perform any turning maneuver when they stop. The measurement innovation for ZU is given as
\begin{equation} \label{measurement_zero}
    \mathbf{z}_{ZU} = [-\mathbf{\omega}_{IMU}, -\mathbf{v}_{eb}^{n}]
\end{equation}
where $\mathbf{\omega}_{IMU}$ represents gyro-rate measurements. 
Similarly, updating the error-state, covariance, and cross-correlated terms follow the same structure through Equations~\ref{kalman_gps}-\ref{correlation_gps}. The Jacobian of the ZU measurement model is described as;
\begin{equation}
\mathbf{H}_{ZU} =
\begin{bmatrix}
\mathbf{0}_{3x3} & \mathbf{0}_{3x3} &\mathbf{0}_{3x3} & \mathbf{0}_{3x3} & \mathbf{-I}_{3x3} \\
\mathbf{0}_{3x3} & \mathbf{-I}_{3x3} & \mathbf{0}_{3x3} & \mathbf{0}_{3x3} & \mathbf{0}_{3x3}
\end{bmatrix}
\end{equation}
The measurement noise covariance matrix for ZU, $\mathbf{R}_{ZU}$, which is a 6-by-6 matrix, and it can be constructed similarly as provided previously in other private updates. 
\subsection{Relative Update}
Relative updates are performed with pairwise ranging whenever two robots are within a specific range and are assumed to occur only when the robots are within a specific separation distance. This relative update model is based on the work in~\citet{luft2016recursive}. In our work, UWB range measurements are used to trigger the relative updates. Whenever relative update is performed, there is one robot that detects the other robot present in the update and receives the state, covariance and cross-correlated terms from it. The decentralized architecture of the error-state EKF allows the robots to decouple their individual states, covariances, and cross-correlated terms to construct a combined covariance matrix as  
\begin{equation} \label{coupling}
    {\mathbf{P}}^{-}_{global}=\left[\begin{array}{cc}
{\mathbf{P}}^{+}_{A} &\mathbf{\Sigma}_{AB}\\
\mathbf{\Sigma}_{AB}^{T} &{\mathbf{P}}^{+}_{B} \end{array}\right], 
\quad \mathbf{\Sigma}_{AB}= \mathbf{\sigma}_{AB}^{+}{\mathbf{\sigma}_{BA}^{+}}^{T}, \quad \mathbf{\Sigma}_{AB}^{T}=\mathbf{\Sigma}_{BA}
\end{equation}

where ${\mathbf{P}}_{global}$ represents the combined covariance matrix, $\mathbf{P}_{A}$ represents the individual covariance matrix for robot $A$, $\mathbf{P}_{B}$ represents the individual covariance matrix for robot $B$, $\Sigma_{AB}$ represents the combined correlated values from robot $A$ and $B$.

The Jacobian of the ranging measurement model is described as
\begin{equation} \label{H_rela}
\mathbf{H}_{range} =\left[\textbf{0}_{1x6}, \left[\frac{(\mathbf{r}_{A}-\mathbf{r}_{B})}{h_{range}}\right],\textbf{0}_{1x12}, \left[\frac{(\mathbf{r}_{B}-\mathbf{r}_{A})}{h_{range}}\right],\textbf{0}_{1x6}\right]^{T}   
\end{equation}
where $\mathbf{r}_{A}$ and $\mathbf{r}_{B}$ represent the 3D position values for robot $A$ and $B$, respectively, such that $\mathbf{r}_{A}=[{x}_{A},{y}_{A},{z}_{A}] $ and $\mathbf{r}_{B}=[{x}_{B},{y}_{B},{z}_{B}] $. 
The non-linear measurement model is described as
\begin{equation} \label{h_rela}
h_{range} = \sqrt{[x_B-x_A]^{2}+[y_B-y_A]^{2}+[z_B-z_A]^{2}}.     
\end{equation}

The Kalman gain for the relative update is calculated as 
\begin{equation} \label{kalman_rela}
\mathbf{K}_{range} = {\mathbf{P}}^{-}_{global} \mathbf{H}_{range}^{T} (\mathbf{H}_{range} {\mathbf{P}}^{-}_{global} \mathbf{H}_{range}^T + \mathbf{R}_{range} )^{-1}.     
\end{equation}

Using the calculated Kalman gain, the error state is updated as
\begin{equation} \label{state_rela}
    \left[\begin{array}{c}
\mathbf{x}_{Aerr}^{+}\\
\mathbf{x}_{Berr}^{+}\end{array}\right]= \left[\begin{array}{c}
\mathbf{x}_{Aerr}^{-}\\
\mathbf{x}_{Berr}^{-}\end{array}\right] + 
\mathbf{K}_{range} \left({z}_{UWB} - \mathbf{H}_{range} \left[\begin{array}{c}
\mathbf{x}_{Aerr}^{-}\\
\mathbf{x}_{Berr}^{-} \end{array}\right]\right)  
\end{equation}
where $\mathbf{x}_{Aerr}$ and $\mathbf{x}_{Berr}$ represent the error state of the robots performing relative update and $z_{UWB}$ represents the ranging measurement. Also, the covariance is updated as
\begin{equation} \label{covariance_rela}
{\mathbf{P}}^{+}_{global}=\left(\mathbf{I}-
\mathbf{K}_{range} \mathbf{H}_{range}\right){\mathbf{P}}^{-}_{global}.
\end{equation}


Once the relative update is completed, the error-state and covariance estimates are sent back to the robot that was detected. The decomposition of the updated correlated values for the robot that performing the update is selected as $\sigma_{AB} \leftarrow \mathbf{I}$ and the robot detecting and performing the update will keep the updated correlated term. The decomposition of the updated correlated values for the robot being detected is selected as $\sigma_{BA} =\mathbf{\Sigma}_{BA}$. The robot receiving the update will get the identity matrix, following the same decomposition used in~\citet{luft2016recursive}.

Lastly, the cross-correlated terms of the robots performing the update with the robots not present in the update is updated with Equation \ref{correlation_rela}.
\begin{equation} \label{correlation_rela}
\mathbf{\sigma}_{AC}^{+}  ={\mathbf{P}}^{-}_{A} {{\mathbf{P}}^{+}_{A}}^{-1} \mathbf{\sigma}_{AC}^{-}  
\end{equation}
where ${{\mathbf{\sigma}}_{AC}} $ represents the cross-correlated terms with the rest of the robots not present in the update.

\section{Experiments}\label{Experiments}
In order to evaluate the algorithm performance, two different experiments are performed. One set in a simulation environment, and another set a real-world environment using wheeled robots. The simulation environment is built in Gazebo/ROS and the real world experiment is performed in a video motion capture facility with instrumented i-Robot Create platforms. In both experiments, the motivation is observing the efficiency of using ZU to reinstate the localization of lost robots. In the following subsections, both experiment settings, robots, and sensors used are explained in detail.   
\subsection{Simulation Experiment Setup}

\begin{figure}[htb] 
{
	{
		\includegraphics[width=0.6\linewidth]{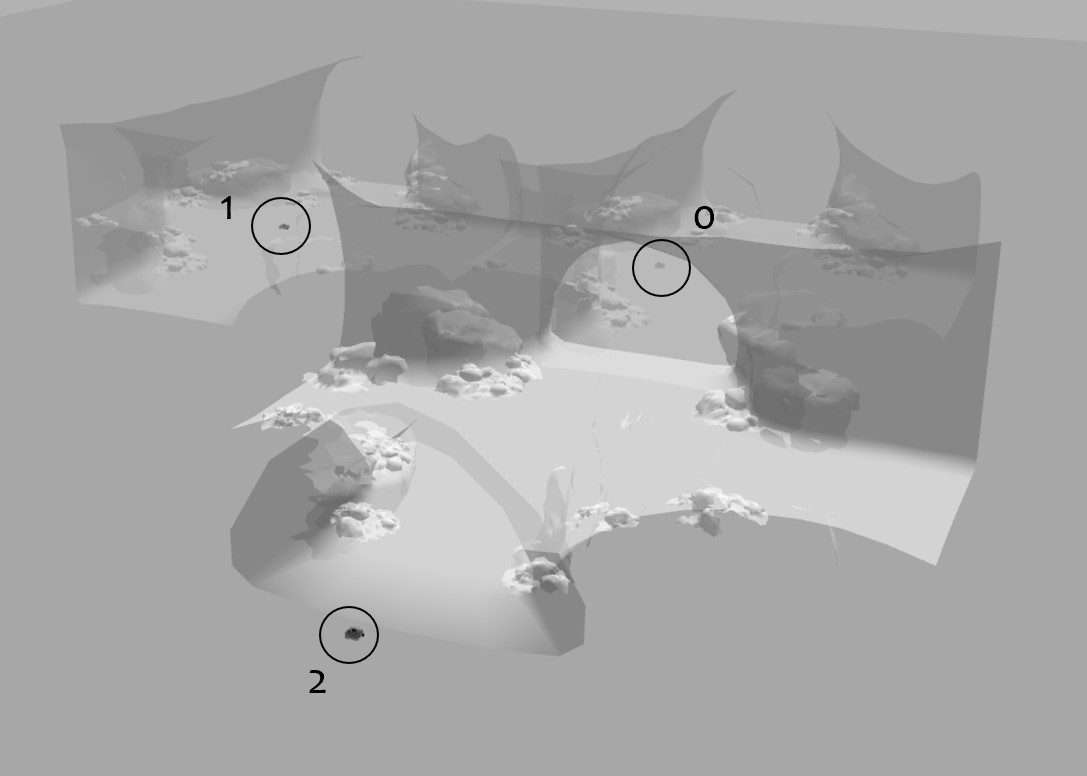}
		\label{global_scenario2}
	}
	{
		\includegraphics[width=0.30\linewidth]{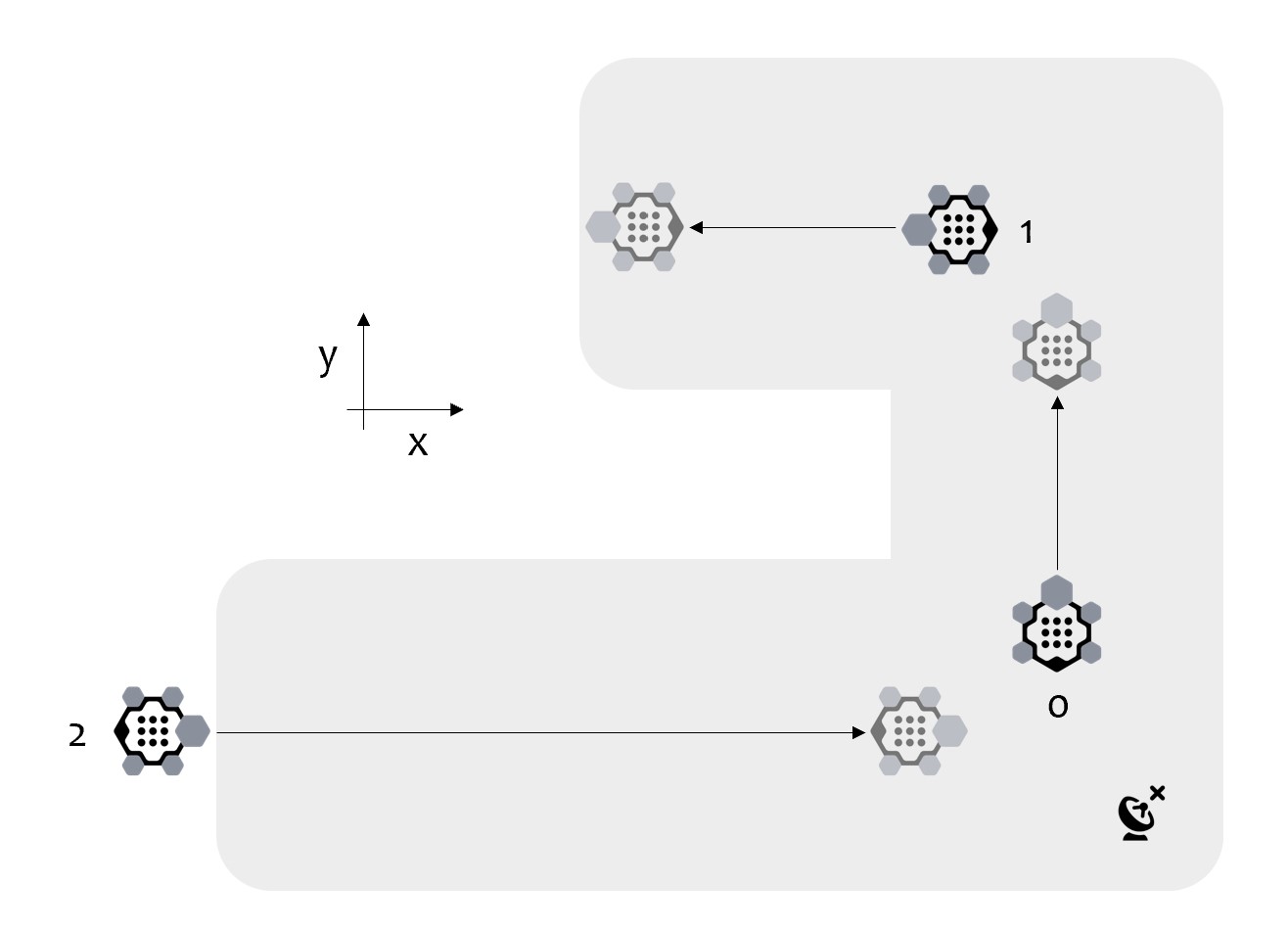}
		\label{local_scenario2}
	}
}    
\caption{Simulation environment with a transparent view. Robot 0, 1 and 2 initial locations in the simulation are shown with black circles. }    
\label{plan_scenario2}
\end{figure}

In the simulation tests, the motivation is reinstating the localization of lost robots in a subterranean environment. It is assumed that two robots became unreliable due to the erroneous localization estimates and they stop moving. Meanwhile, another robot is deployed for a mission to restore the localization estimates of the lost robots. The scenario starts with a robot, named Robot 2, entering a cave just after receiving reliable GNSS signals. The other robots (i.e., Robot 0 and Robot 1) are unable to use either visual perception or GNSS for localization purposes due to the poor lighting conditions of the cave and the blockage of the GNSS signals. Robot 2 traverses a straight line on the $x$-axis using its IMU sensor and wheel encoders to estimate its position while leveraging ZU under stationary conditions. Robots can perform relative update pairwise, whenever they are in a close proximity to each other which is set by a predetermined threshold. This motivating case is illustrated in Fig.~\ref{plan_scenario2}, where the grey area in the bird-eye view (right sub-figure) represents the GNSS-denied subterranean environment.

The tests in the simulation environment include three simulated TurtleBot3~\citep{amsters2019turtlebot} with the sensor models for wheel encoders, {IMU}, {GNSS}, and {UWB}. Additive white Gaussian noise values are added to the default outputs of the provided sensors to simulate the effect of random process. The rate of the sensors and the noise characteristics of the used sensors are provided in Table~\ref{gazebo_noise}.  

\begin{table} [htb]
\centering
\footnotesize
\caption{TurtleBot3 Sensor Parameters}
\label{gazebo_noise}
\centering
\begin{tabular}{@{}llccccc@{}}
\hline
Sensor& Measurement & Noise 1$\sigma$& Rate\\ 
\hline
\multirow{2}{*}{\textbf{{IMU}}}
 & Acceleration      & 0.001 $m/s^{2}$	     & 50 Hz\\
 & Gyro Rate         & 0.001 $rad/s$        & \\ 
\textbf{{Encoder}}
 & Velocity          & 0.01 $m/s$ 	         & 30 Hz\\
 
\multirow{2}{*}{\textbf{{GNSS}}}
 & Position          & 0.1 $m$	             & 1 Hz\\
 & Velocity          & 0.02 $m/s$           & \\ 

\textbf{{UWB}}
& Range             & 0.05 $m$             & 1 Hz\\
\hline
\end{tabular}
\end{table}

In simulation tests, communication and ranging measurements are limited to distances shorter than 2.5 meters to simulate an obstructed line of sight capability and all robots move with a velocity of 0.2 m/s, and the simulation environment is assumed to be a flat region, whereas the algorithm is able to provide 3D position estimation. The Robot 2 is able to stop autonomously to perform ZU whenever any of the diagonal elements of the position error covariance reach the predetermined threshold of 5~m$^2$. Given the sensor parameters and scenario settings, this threshold is set based on engineering judgment with respect to the localization reliability and traversal rate. For example, if the robot is not able to reduce the position error covariance lower than the threshold, then the robot starts performing periodic {ZU} to keep the localization performance as reliable as possible. In this respect, the decision of which ZU scheme to utilize depends on several factors, as elaborated in our previous work~\citep{edu2022pseudo}. Using autonomous stopping heuristic is advantageous over periodic stopping to increase the traversal rate, which allows robots to make use of the ZU when the estimated position error covariance exceeds a certain threshold. This is particularly useful in mitigating the effects of IMU drift while minimizing interruptions to the robot's mission. However, the robot's ability to estimate its position error covariance reliably may diminish due to prolonged external aiding outages, sensor errors, or environmental conditions. In such cases, performing ZU at fixed time intervals irrespective of the estimated error may provide consistent and regular correction to the IMU drift with a cost of traversal rate decreasing. For these reasons, we opted to use the autonomous stopping criteria first, and then switched to the periodic stopping criteria. This allows us to take advantage of the benefits of both schemes while minimizing the drawbacks.

The stopping condition is verified through the robot’s wheel encoders and velocity command output after receiving a stop command. Then, after verifying the robot is completely stopped, the robot waits 0.5 seconds to utilize zero updates. This stopping duration is conservatively selected based on engineering judgment and considerations of reliable ROS message delivery during the stopping phase. After waiting time, the robot resumes moving. The actions the robot takes, from verifying its stopping condition to resuming movement, are dictated by a Boolean-based state machine framework. The framework models the behavior of a system using a finite number of states, each associated with a set of actions. The system then transitions between these states based on certain conditions.

In the context of the rover stopping mechanism, it is important to mention potential challenges such as false positives and negatives. If a robot mistakenly assumes it has stopped (false positive), this could introduce errors, while if it fails to recognize a stop (false negative), potential drift correction opportunities could be missed, leading to less accurate localization. To handle these challenges, the state machine framework uses Boolean indicator flags in ROS to determine when the robot should stop and start moving again based on the feedback from the wheel encoders and velocity command output. The threshold parameters for autonomous ZU, the duration of movement during periodic ZU, and the stopping duration can be easily customized in the provided code based on the mission scenario and robot constraints.

\subsection{Real-World Experiment Setup}

In the real-world experiments, using a similar motivation with simulation experiments, the aim is restoring the localization of the lost robots. The difference from the simulation setup, the lost robots do not stay in stationary conditions before their first relative update. The real-world experiments are performed in a 3x3 meters room with video motion capture system. Also, VICON is used to simulate a GNSS update in the indoor test settings for initializing the Robot 2 pose and generating the truth solution for all robots. Robots are assumed that they cannot use visual odometry due to poor
lighting conditions and lack of sufficient features in the environment; however, robots can use their {IMU}, wheel encoders, and {UWB} sensors.

\begin{figure}[t!] 
{
	{
		\includegraphics[width=0.75\linewidth]{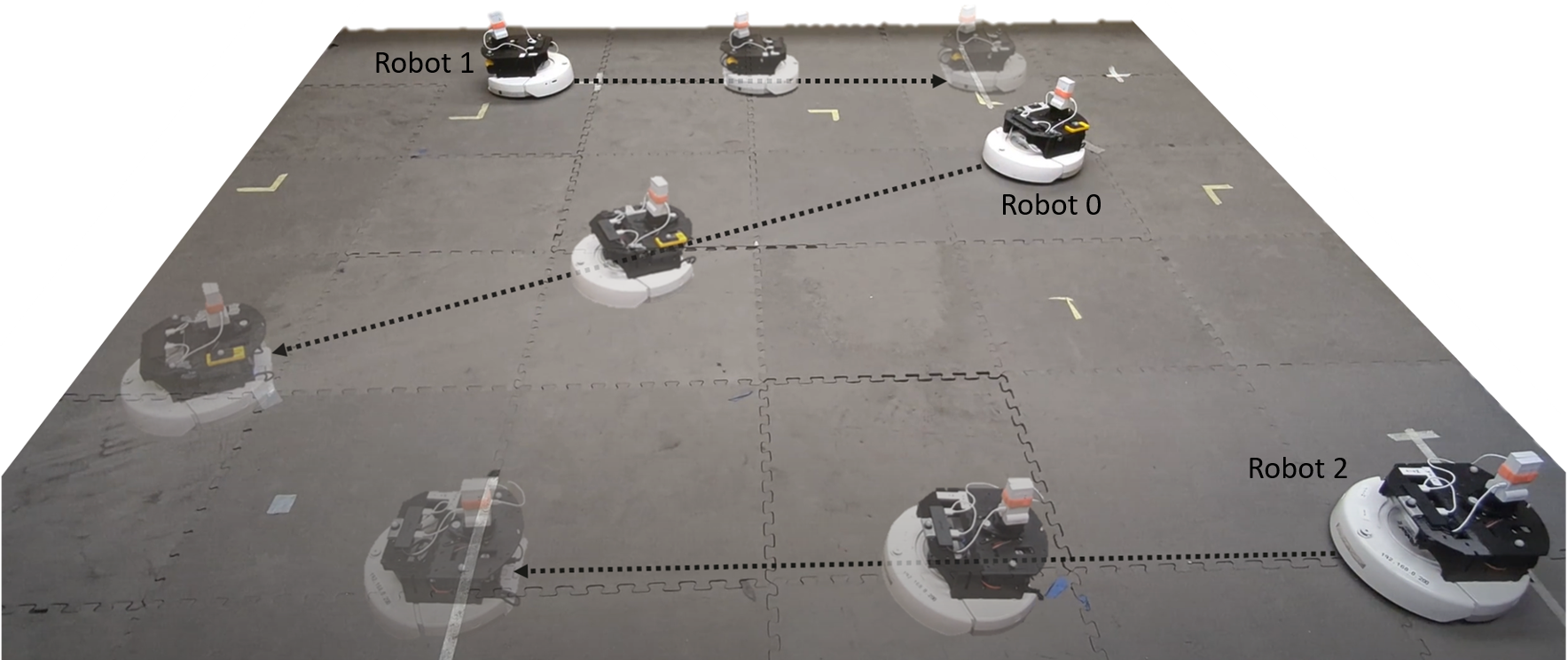}
		\label{global_scenario2}
	}\hfill
	{
		\includegraphics[width=0.19\linewidth]{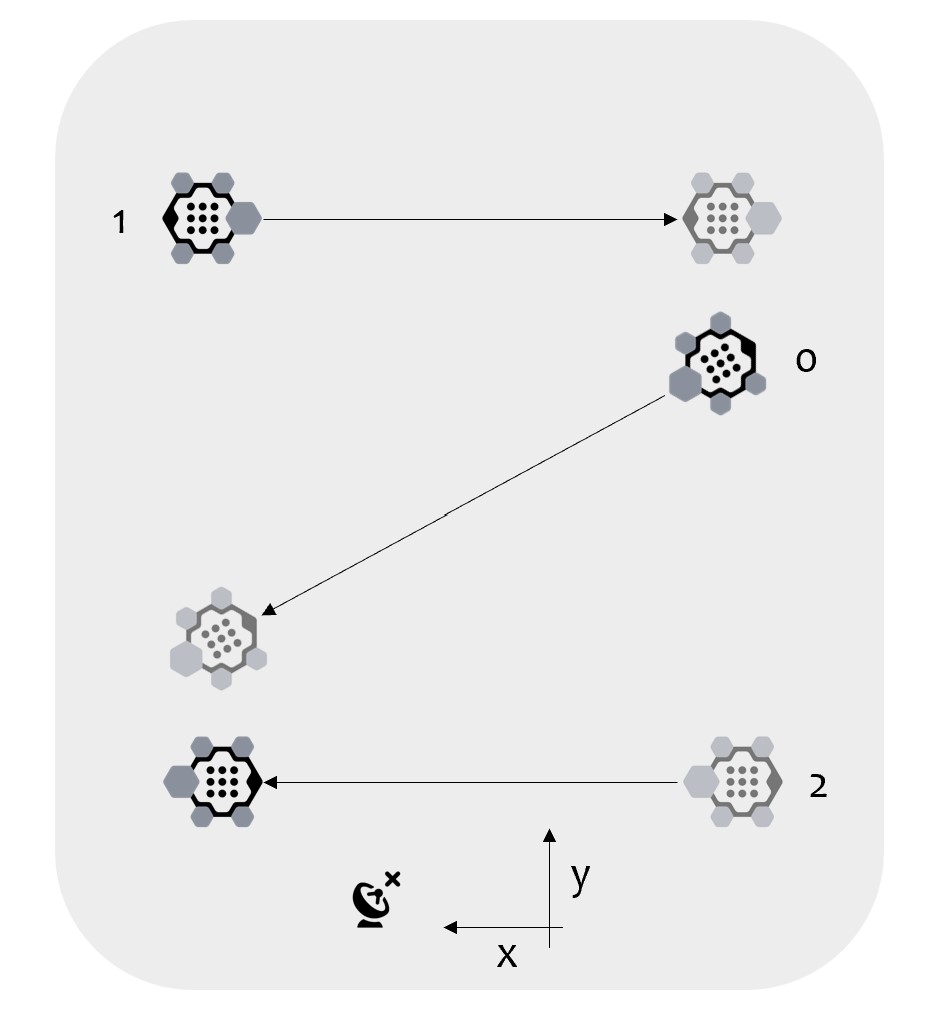}
		\label{local_scenario2}
	}
}    
\caption{Real-world experiment setting.}    
\label{hardware}
\end{figure}

In these experiments, similar to the cave simulation experiments, it is assumed that Robot 2 starts moving with a known and accurate localization whereas the Robot 0 and 1 are lost during their patrols in the area. Robot 2 patrols the $x$-axis, Robot 0 makes a diagonal movement across the room, and Robot 1 patrols the $x$-axis with an offset in $y$-axis to make sure that Robot 2 or Robot 1 can only perform relative update with Robot 0 (i.e., Robot 2 and Robot 1 cannot detect or communicate with each other). With this setup, the localization performance of Robot 1 depends on the localization performance of Robot 0, and the performance change on Robot 2 using or not using ZU will affect the entire system. Robots have a velocity of 0.2 m/s and the detection limit for robots with range is set to 1 meter due to dimension restrictions. The test environment is given in Fig.~\ref{hardware}.

\begin{table} [h!]
\centering
\footnotesize
\begin{threeparttable}
\caption{Inertial Measurement Unit Specifications}
\label{tab:imuspec}
\centering
\begin{tabular}{@{}lccccccccc@{}}
\hline
 & \multicolumn{3}{c}{Gyroscope}   &\multicolumn{3}{c}{Accelerometer}&\multicolumn{2}{c}{Physical Specs }\\
IMU & \scriptsize{Range}& \scriptsize{In-run bias }& \scriptsize{ARW\tnote{*} } & \scriptsize{Range}& \scriptsize{In-run bias }& \scriptsize{VRW\tnote{*}}  &\scriptsize{Dimensions}& \scriptsize{Weight}
\\ 

& \scriptsize{(deg/sec)}& \scriptsize{(deg/hr)}& \scriptsize{(deg/$\sqrt{hr}$)} & \scriptsize{(g)}& \scriptsize{(mg)}& \scriptsize{(m/s/$\sqrt{hr}$)} &\scriptsize{ (mm)} & \scriptsize{(g)}
\\ 
\hline\hline
ADIS16405BMLZ   &$\pm$350 &25.2 &2.0 &$\pm$18 &0.2 & 0.2 & 32 x 23 x 23 & 16 \\

\hline
\end{tabular}
\begin{tablenotes}
\item[*] ARW: Angular Random Walk, VRW: Velocity Random Walk
\end{tablenotes}
\end{threeparttable}
\end{table}
The i-Robot Create robots are used in the real-world experiments. To share the information between robots and record the VICON solution data, the robots and the VICON system are connected to the same network via Wi-Fi. The specification and noise characteristics of the {IMU}, ADIS16405, are given in Table~\ref{tab:imuspec}. {UWB} sensor, DWM1001-DEV, measurement specification are given as $\pm 15$~cm and $\pm 30$~cm for 2D and 3D noise, respectively. In these experiments, we followed the same autonomous stopping heuristic previously described for Robot 2 in the simulation experiments. In this case, the threshold for stopping is predetermined to 2 m$^2$ to observe the effectiveness of ZU before the first relative update between robots due to dimension restrictions. Note that, while the environment used in our experiments was rigid, flat, and benign, with well-behaved dynamics on the i-Robot Create and slow velocity inputs, we acknowledge that in harsher environments or with excessive slippage, additional indicators may be needed to ensure the robot stops completely.






\section{Evaluation}
\label{Evaluation}

\subsection {Simulation Results}

\begin{figure}[t!] 

	{
		\includegraphics[width=0.33\textwidth]{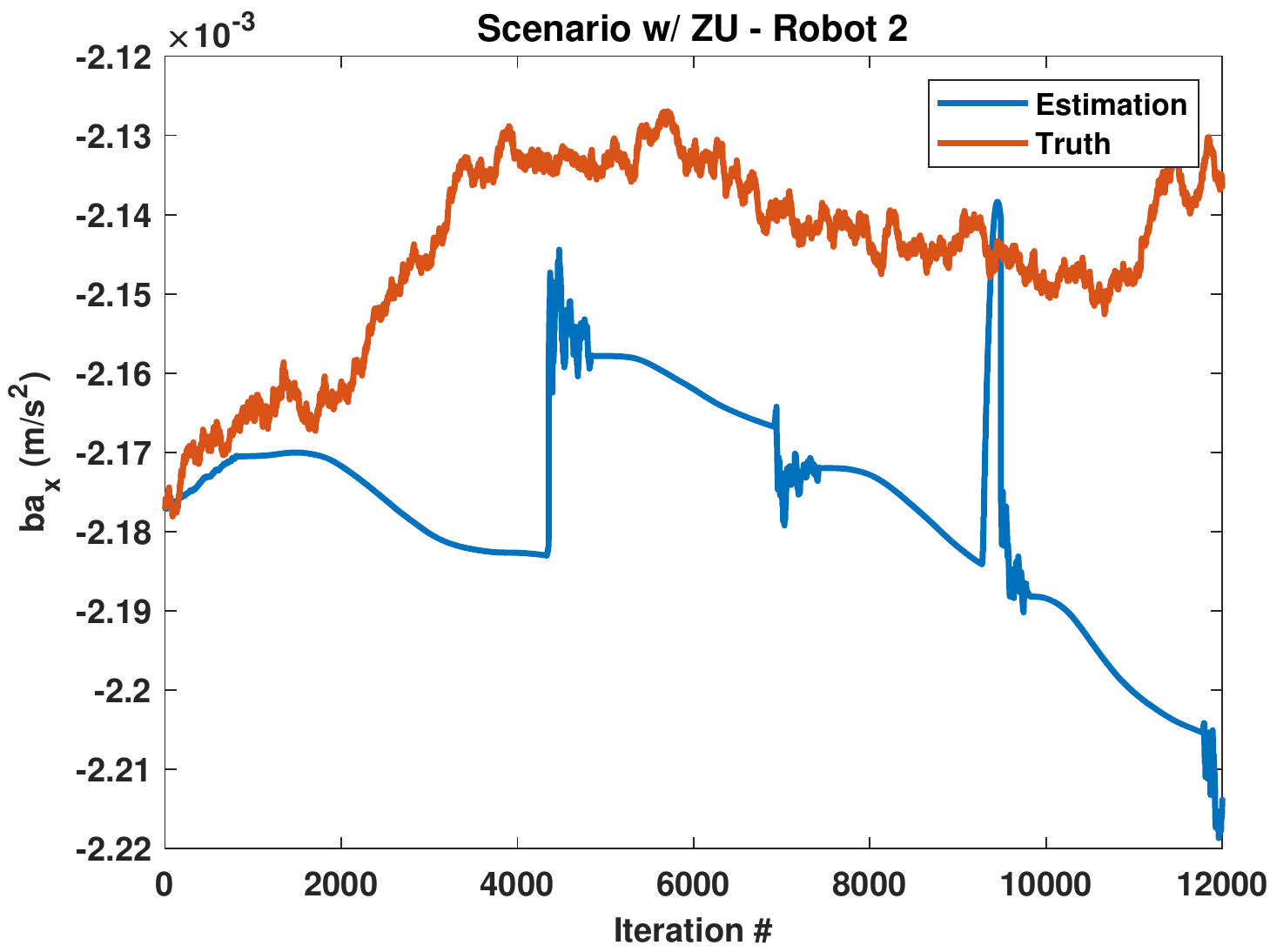}
		\label{fig:baxzu}
	}
	{
		\includegraphics[width=0.33\textwidth]{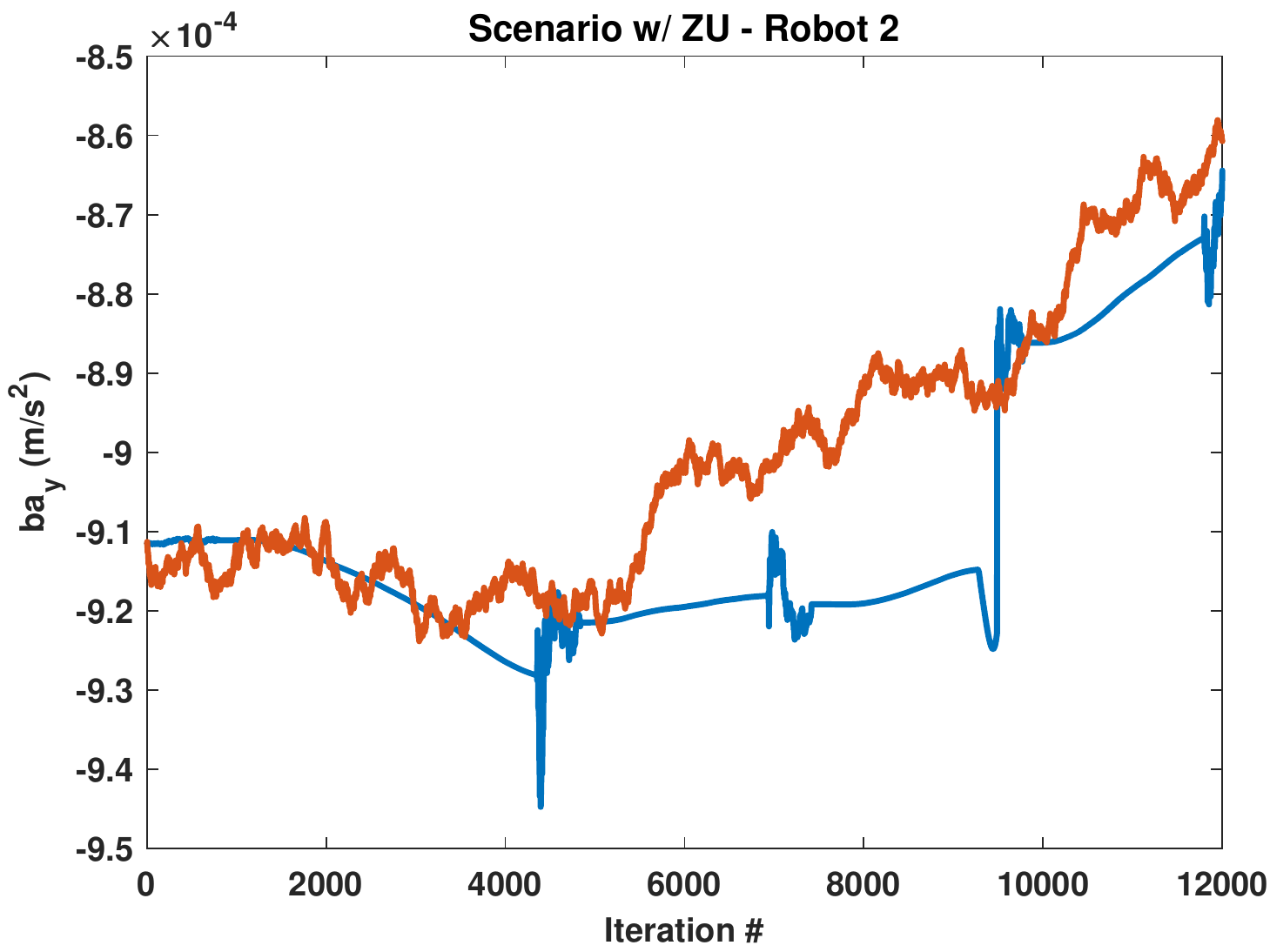}
		\label{fig:bayzu}
	}
	{
		\includegraphics[width=0.33\textwidth]{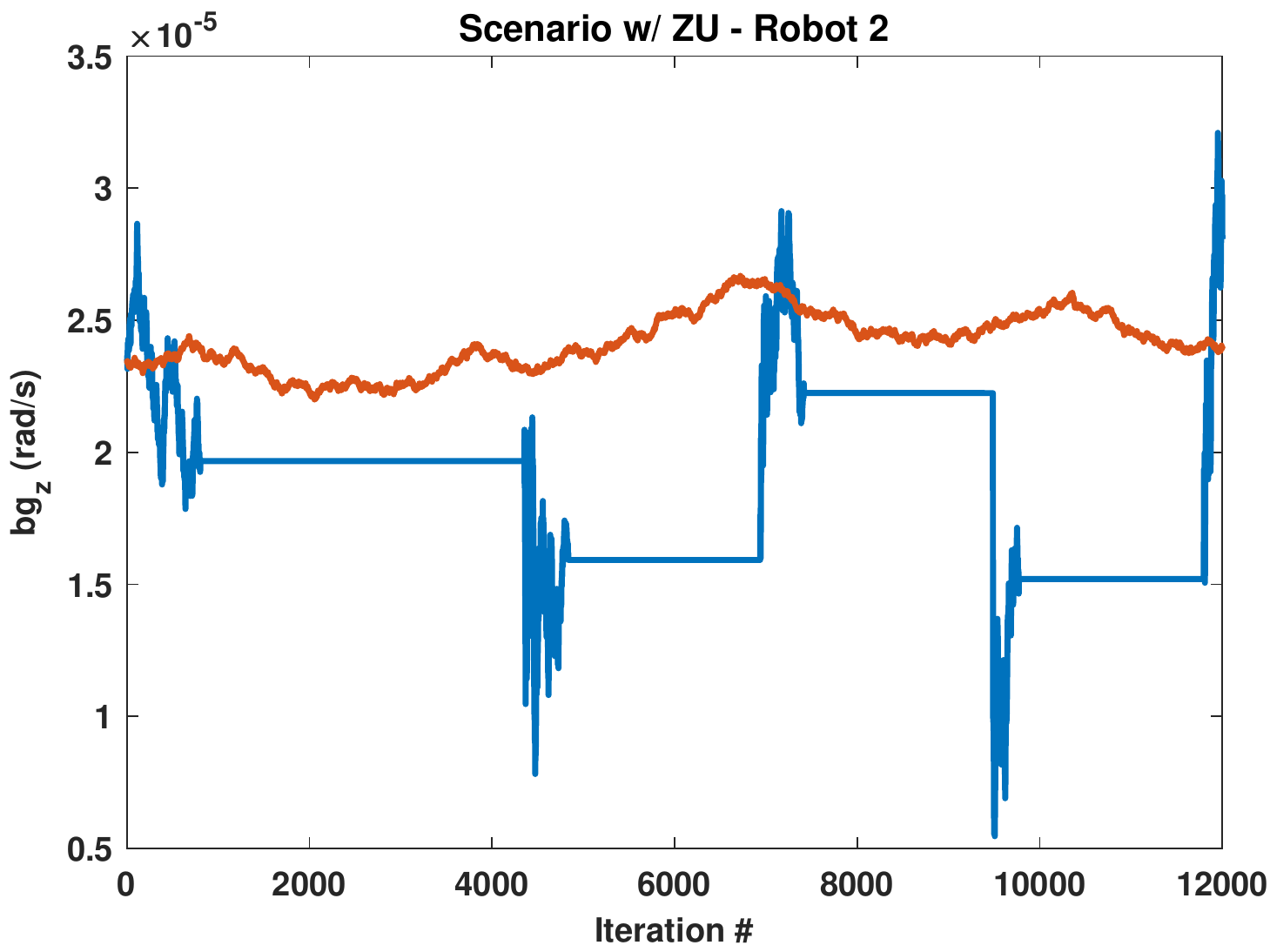}
		\label{fig:bgzzu}
	}
	
	{
		\includegraphics[width=0.33\textwidth]{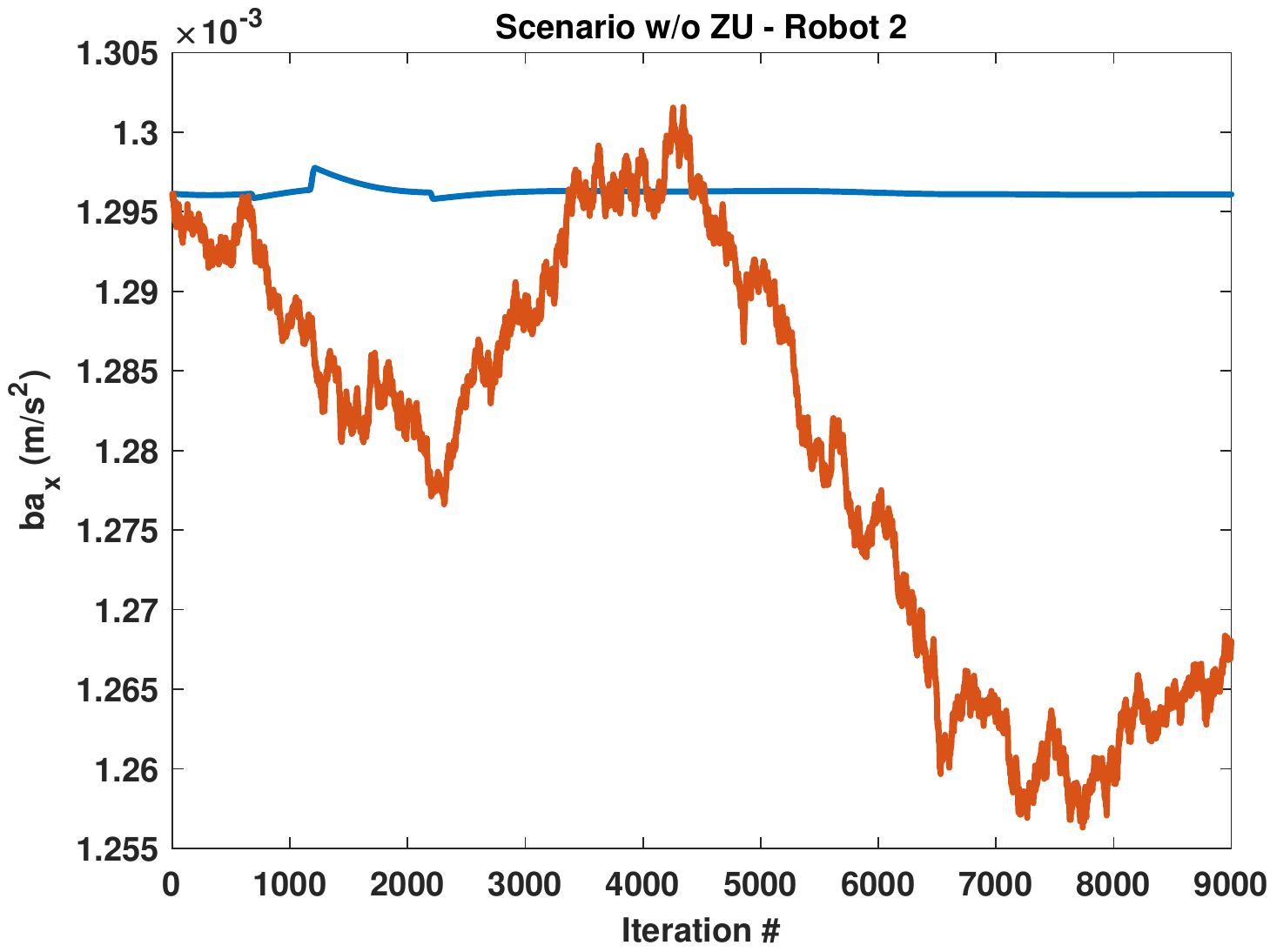}
		\label{fig:baxnozu}
	}
	{
		\includegraphics[width=0.33\textwidth]{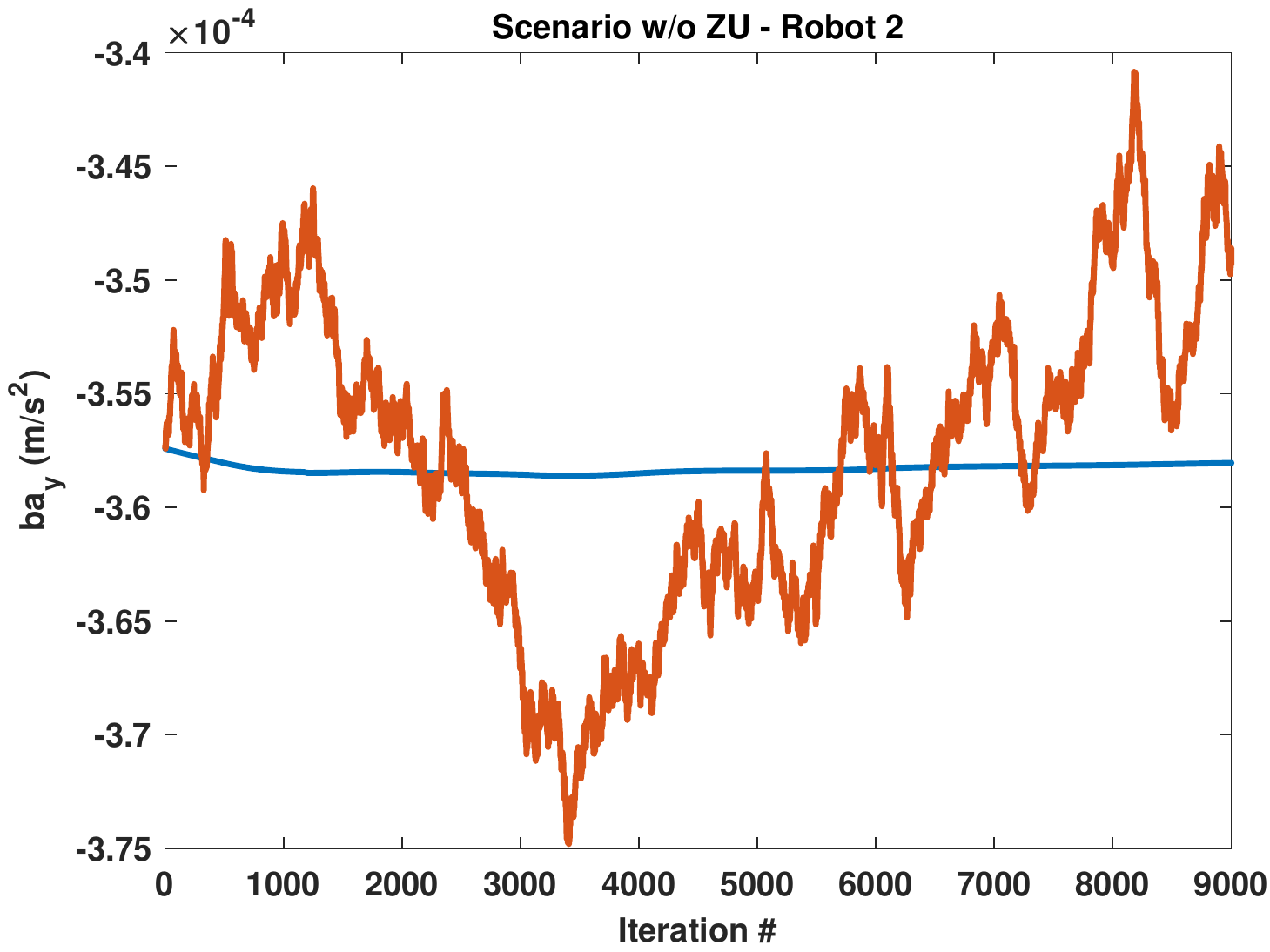}
		\label{fig:baynozu}
	}
	{
		\includegraphics[width=0.33\textwidth]{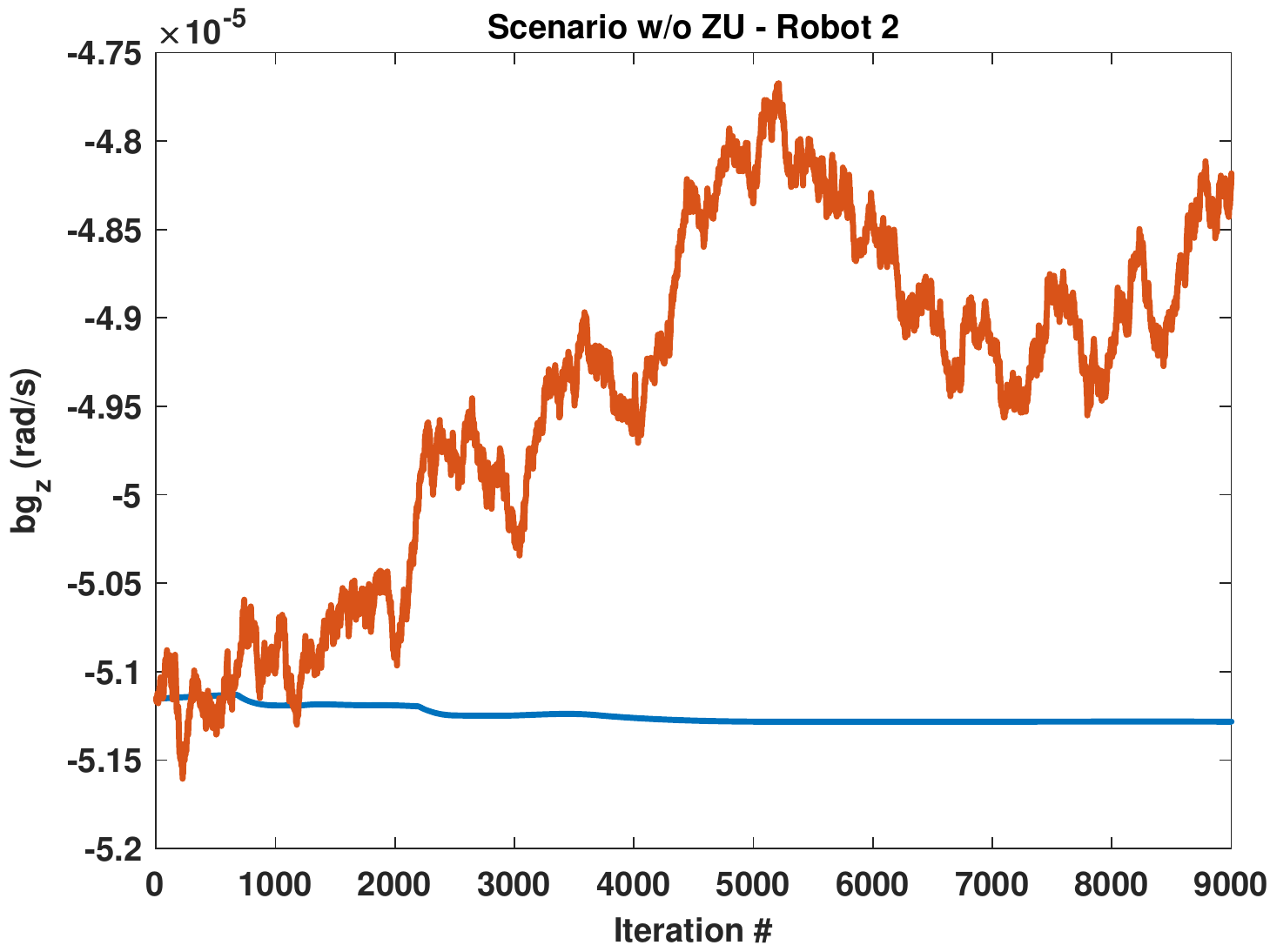}
		\label{fig:bgznozu}
	}
\caption{Example scenarios for estimated bias comparisons with respect to the truth generated through simulation. Scenario w/ ZU show the bias estimation and simulated IMU sensor bias from the Gazebo model (i.e., truth) for Robot 2 when performing ZU during its traversal. Scenario w/o ZU presents the case when Robot 2 does not perform ZU. The x-axis is labeled as "iteration~$\#$", which corresponds to the IMU measurement updates occurring at 50Hz, essentially representing time steps relative to the IMU measurements. The choice of "iteration~$\#$" instead of "time" highlights the dependency of IMU bias calibration on the number of iterations rather than the actual duration. It can be observed that, in Scenario w/ ZU, the bias estimation attempts to converge towards the true bias values. In contrast, in the scenario w/o ZU, the bias estimation changes slightly, but the effect of ZU is not present. }    
\label{fig:biases}
\end{figure}

For the simulation experiments, we first analyze the localization performance of Robot 2 and whether ZU is being leveraged through its traversal or not. The position errors for both cases and the average improvement for each metric are shown in Table~\ref{tab:comparison_sim}. Applying ZU can bound the velocity error, calibrate IMU sensor biases (see Fig.~\ref{fig:biases}), and limit the rate of INS localization error growth \cite{grovebook}.

In environments where robots are not able to use their visual sensors or rely on GNSS signals, using ZU provides a significant improvement and shows the effectiveness of keeping the localization estimation reliable in all axes as in Table~\ref{tab:comparison_sim}.

\begin{figure}[t!] 
{
	{
		\includegraphics[width=0.33\textwidth]{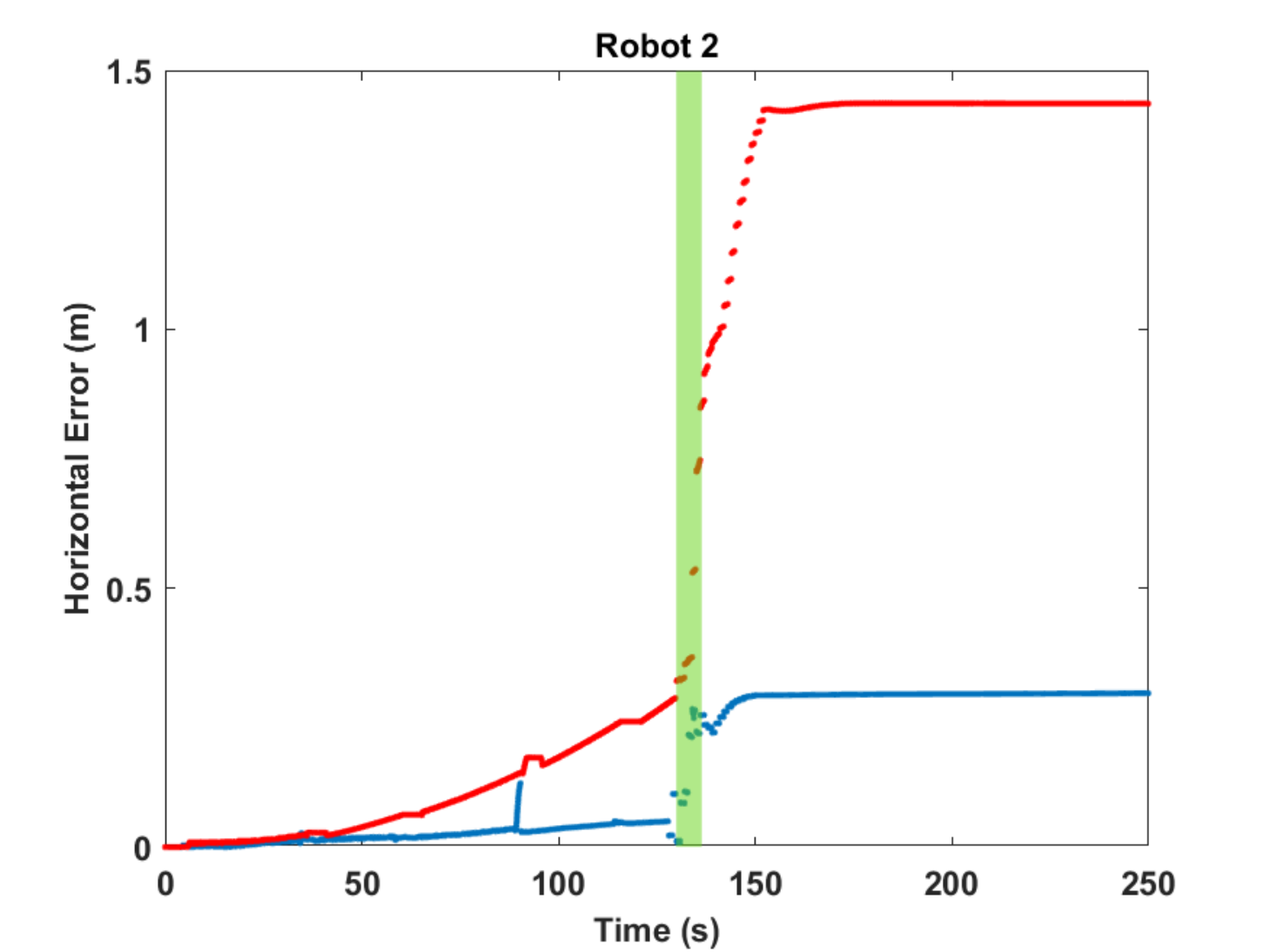}
		\label{fig:robot2}
	}
	{
		\includegraphics[width=0.33\textwidth]{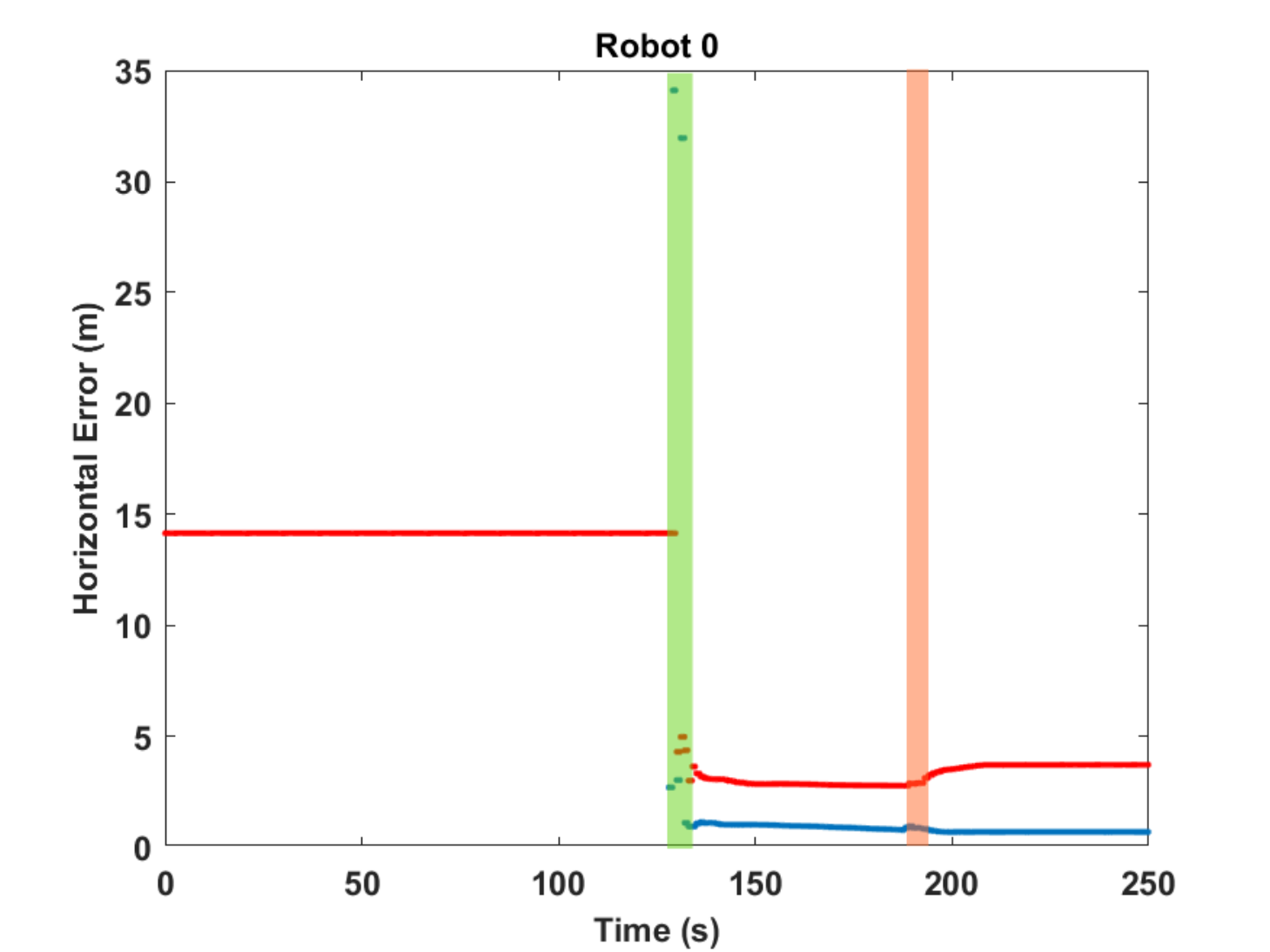}
		\label{fig:robot0}
	}
	{
		\includegraphics[width=0.33\textwidth]{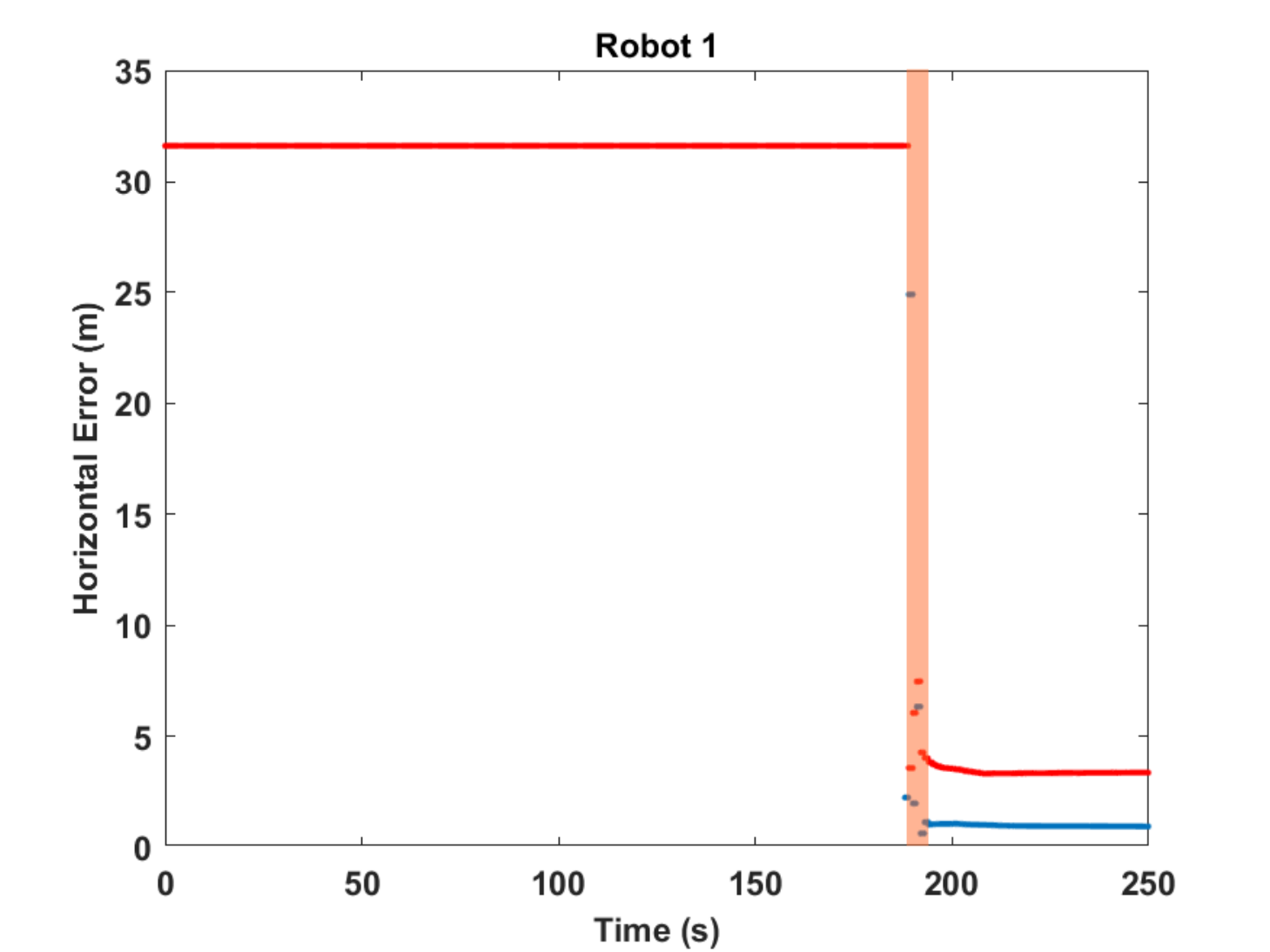}
		\label{fig:robot1}
	}
}    
\caption{Horizontal error estimation performance comparison for the cases where ZU is used versus not used by Robot 2 in simulation experiment (Test 1). The red dots represent the estimation error when ZU is not used, the blue dots represent the estimation error when ZU is used. The green shaded areas show the approximate duration when Robot 2 performs a relative update with Robot 0, and the orange shaded area show when Robot 0 performs a relative update with Robot 1. }    
\label{fig:simulation_1}
\end{figure}

\begin{table} [htb]
\centering
\footnotesize
\begin{threeparttable}
\caption{Localization estimation comparison of Robot 2 in simulation experiments}
\label{tab:comparison_sim}
\centering
\begin{tabular}{@{}clccccc|ccccc|cc@{}}
\hline
& & \multicolumn{5}{c}{w/o ZU} & \multicolumn{5}{c}{ZU}\\ 
\hline
&\textbf{Robot 2 }& \scriptsize{T1}& \scriptsize{T2}& \scriptsize{T3}& \scriptsize{T4}& \scriptsize{T5}&\scriptsize{T1}& \scriptsize{T2}& \scriptsize{T3}& \scriptsize{T4}& \scriptsize{T5}& \scriptsize{Improvement*}\\
\hline
\multirow{6}{*}{\rotatebox[origin=c]{90}{{\centering Error (m)}}}&
$\text{Max}_{E}$      & 0.53 & 0.27 & 0.55 & 0.36 & 15.71 & 0.12 & 0.13 & 0.18 & 0.31 & 0.18 & 94.66\% \\
&$\text{Max}_{N}$     & 1.34 & 1.02 & 0.74 & 0.48 & 4.38 & 0.29 & 0.34 & 0.27 & 0.26 & 0.26 & 82.13\% \\
&$\text{Max}_{U}$     & 403.06 & 402.95 & 368.89 & 259.97 & 303.31& 8.52 & 7.29 & 7.38 & 4.91 & 6.15 & 98.03\% \\
&$\text{RMSE}_{E}$    & 0.36 & 0.19 & 0.37 & 0.25 & 8.94 & 0.03 & 0.08 & 0.03 & 0.15 & 0.04 & 96.65\% \\
&$\text{RMSE}_{N}$    & 0.90 & 0.71 & 0.49 & 0.17 & 2.87 & 0.20 & 0.19 & 0.18 & 0.15 & 0.06 & 84.70 \% \\
&$\text{RMSE}_{U}$    & 179.92 & 182.87 & 165.57 & 116.63 & 135.35 & 3.42 & 3.47 & 3.18 & 2.19 & 2.98 & 98.05\% \\
\hline
\multirow{4}{*}{\rotatebox[origin=c]{90}{{\centering 3D Err (m)}}}&
Median      & 97.76 & 112.22 & 93.58 & 67.19 & 73.38  & 3.77 & 4.08 & 3.46 & 2.65 & 3.44 & 96.08\% \\
&Max         & 403.06 & 402.95 & 368.89 & 259.97 & 303.58 &8.52 & 7.30 & 7.38 & 4.92 & 6.15 & 98.03\% \\
&STD         & 120.53 & 119.48 & 109.94 & 77.37 & 91.22 & 1.32 & 1.17 & 1.17 & 0.79 & 0.95 &98.96\% \\
&RMSE        & 179.92 & 182.87 & 165.58 & 116.63 & 135.67 & 3.43 & 3.47 & 3.18 & 2.20 & 2.98 & 98.04\% \\
\hline

\end{tabular}
\begin{tablenotes}
\item *Improvement calculated by using the mean values of 5 tests for each metric ($100 \times(\Bar{\text{w/o ZU}}-\Bar{\text{w/ ZU}})/\Bar{\text{w/o ZU}}$).
\end{tablenotes}
\end{threeparttable}
\end{table}
Leveraging ZU only in a single robot can benefit the overall performance of the multi-robot system in a cooperative localization framework as shown in Table~\ref{tab:r1r2relupd} and in Fig.~\ref{fig:simulation_1}. Table~\ref{tab:r1r2relupd} shows the total horizontal error correction of the lost robots and the improvement percentage when Robot 2 utilizes ZU or not (w/o ZU). The correction is calculated by using the initial and the end of mission horizontal errors for each robot. The approximate time of each pairwise relative update between robots and the horizontal errors are given in Fig.~\ref{fig:simulation_1} where the red dots represent the horizontal errors of each robot in the case that Robot 2 does not use ZU, while the blue dots represent the horizontal error when the Robot 2 leverages ZU. Green and orange rectangles indicate the Robot~2 - Robot~0 and Robot~0 - Robot~1 relative updates, respectively.  In the beginning of the test, Robot 2 is assumed to start moving with accurate position estimation. Robot 0 stays still with a high localization estimation error until it communicates with the Robot 2. Similar to the Robot 0, Robot 1 has high localization error and stays in stationary condition. During Robot 2's traversal, it performs autonomous ZU to keep the estimation reliable. As seen in Fig.~\ref{fig:simulation_1}, using ZU in Robot 2 allows for more positioning correction for Robot 0. Once the relative update between Robot 2 and Robot 0 is done, Robot 0 traverses through the $y$-axis to Robot 1. After the relative update between Robot 0, Robot 1 gains a better localization estimation and start moving through the $x$-axis. Note that only Robot 2 performs ZU, Robot 1 and Robot 0 do not leverage ZU; however, Robot 0 corrects more error when Robot 2 uses ZU, which also benefits to the Robot 1 when it performs relative update with Robot 0. In other means, the benefit of utilizing ZU in Robot 2 is carried by Robot 0 to Robot 1, even Robot 2 and Robot 1 do not share information.      

\begin{table} [t!]
\centering
\footnotesize
\begin{threeparttable}
\caption{Robot 1 and Robot 0 relative update performance analysis in simulation }
\label{tab:r1r2relupd}
\centering
\begin{tabular}{@{}clcc|cc|cc|cc@{}}
\hline
& & \multicolumn{4}{c}{Robot 1 Initial Error: 31.62} & \multicolumn{4}{c}{Robot 0 Initial Error: 14.14}\\ 
\hline
& & \multicolumn{2}{c}{w/o ZU used in Robot 2} & \multicolumn{2}{c}{w/ ZU used in Robot 2}& \multicolumn{2}{c}{w/o ZU used in Robot 2} & \multicolumn{2}{c}{w/ ZU used in Robot 2}\\ 
\hline
&\textbf{ }& \footnotesize{Correction (m)}& \footnotesize{Improvement (\%)}& \footnotesize{Correction (m)}& \footnotesize{Improvement (\%)}& \footnotesize{Correction (m)}& \footnotesize{Improvement (\%)}& \footnotesize{Correction (m)}& \footnotesize{Improvement (\%)}\\
\hline
\multirow{5}{*}{\rotatebox[origin=c]{90}{{\centering Horiz. Error }}}&
T1      & 28.27 & 89.39 \% & 30.70 & 97.07 \% & 10.43 & 73.75 \% & 13.46 & 95.17 \%  \\
&T2     & 30.09 & 95.14 \% & 30.90 & 97.72 \% & 11.72 & 82.92 \% & 12.94 & 91.53 \%  \\
&T3     & 28.13 & 88.96 \% & 30.02 & 94.93 \% & 11.00 & 77.76 \% & 12.62 & 89.27 \% \\
&T4    &  21.23 & 67.13 \% & 29.55 & 93.45 \% & 3.52 & 24.93 \% & 11.99 & 84.81 \%  \\
&T5    &20.44 & 64.63 \% & 30.48 & 96.40 \% & 2.62 & 18.50 \% & 12.72 & 89.99 \% \\
\hline
\end{tabular}
*Improvement is calculated by dividing the correction by the initial error of the robots.
\end{threeparttable}
\end{table}

\begin{figure}[t!] 

	{
		\includegraphics[width=0.33\textwidth]{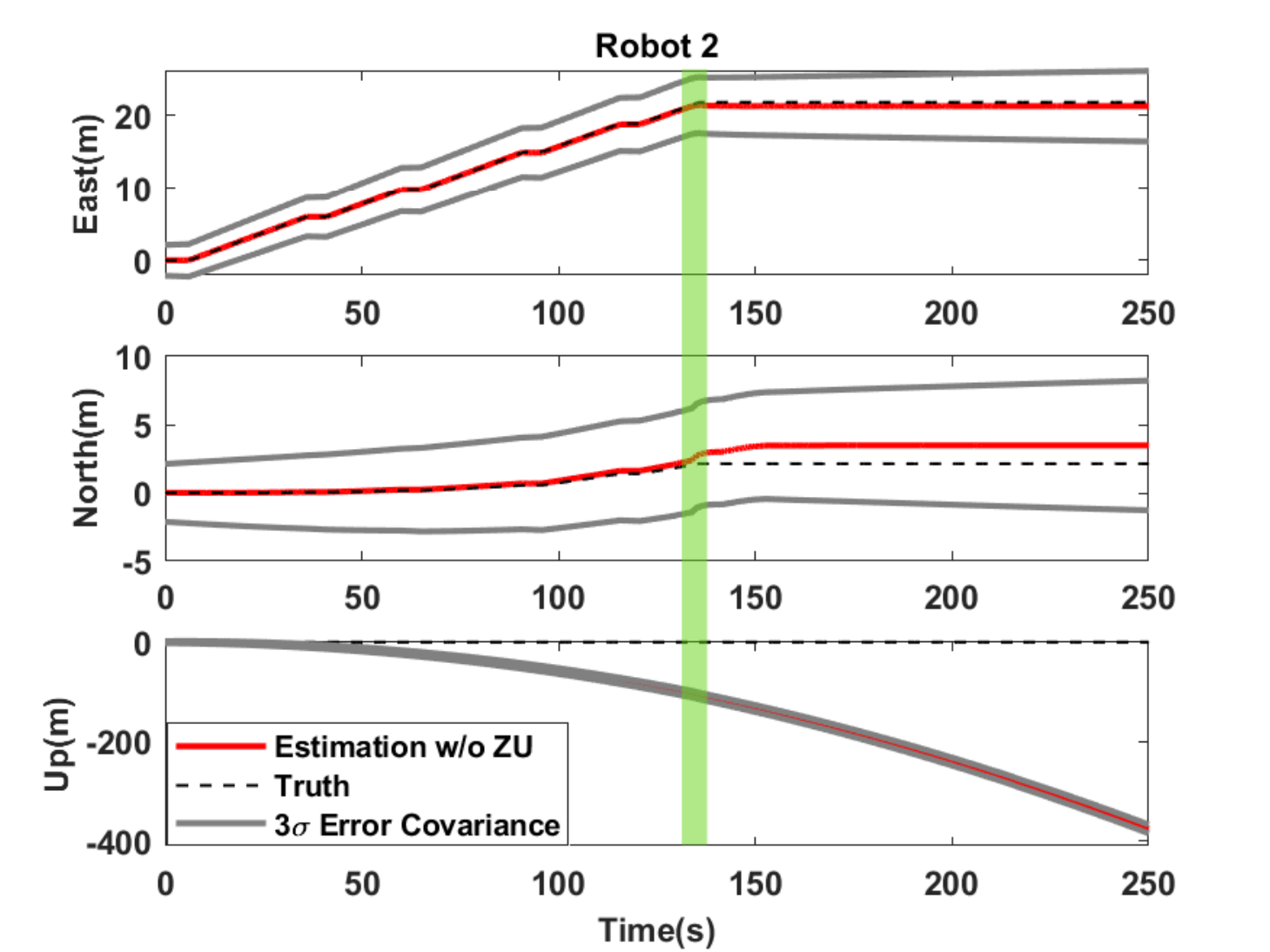}
		\label{fig:robot2_sim}
	}
	{
		\includegraphics[width=0.33\textwidth]{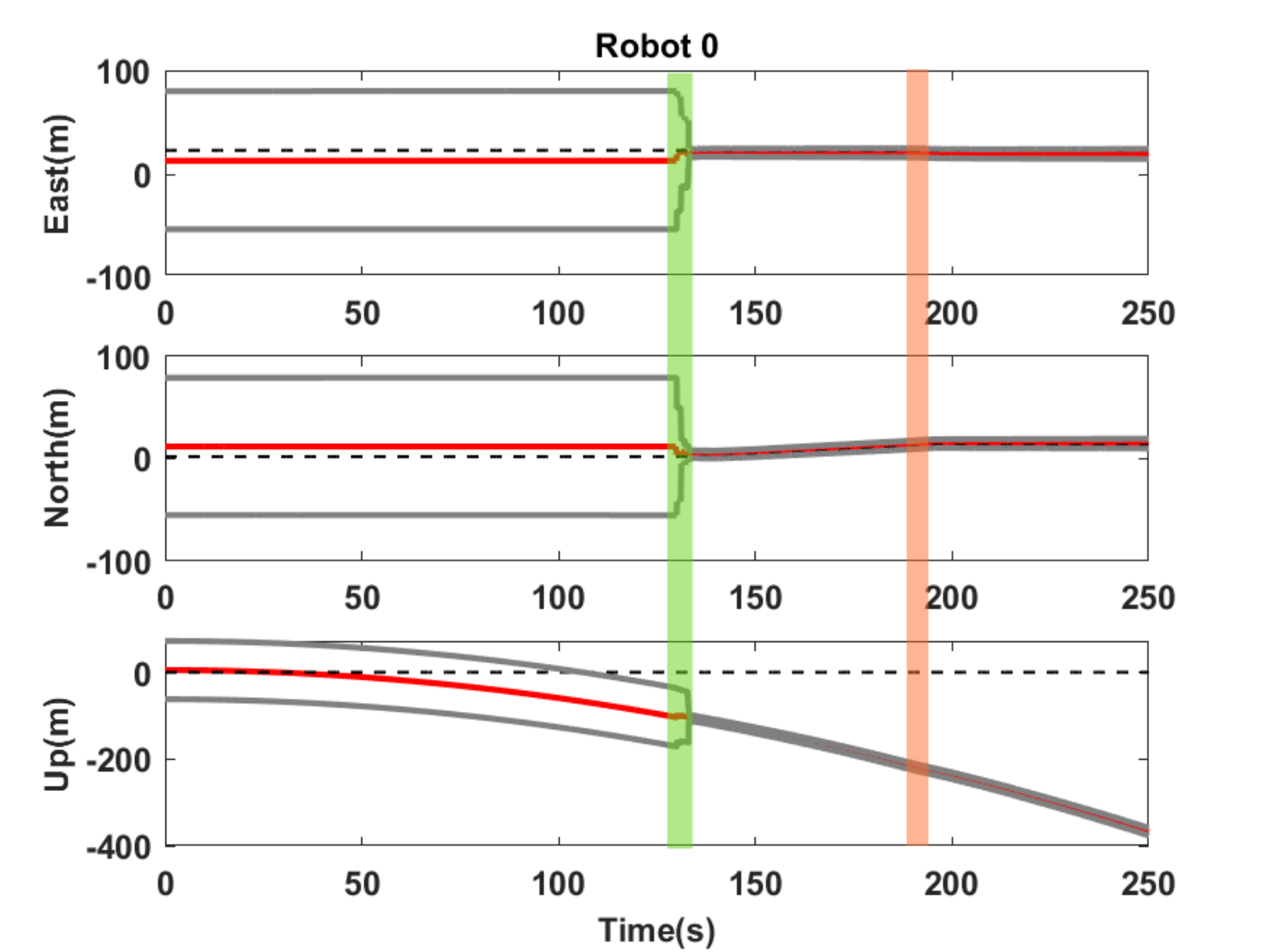}
		\label{fig:robot0_sim}
	}
	{
		\includegraphics[width=0.33\textwidth]{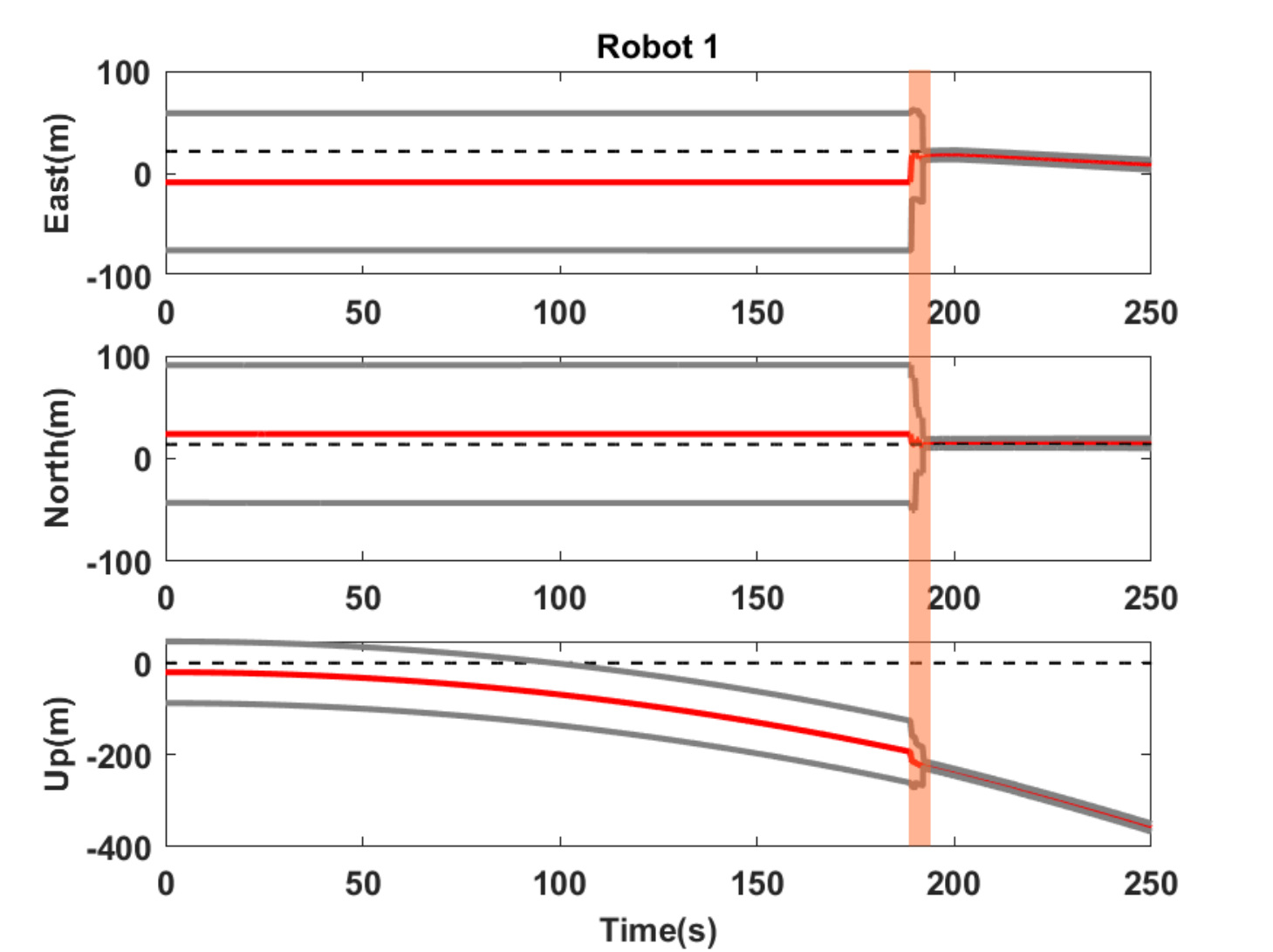}
		\label{fig:robot1_sim}
	}
	
	{
		\includegraphics[width=0.33\textwidth]{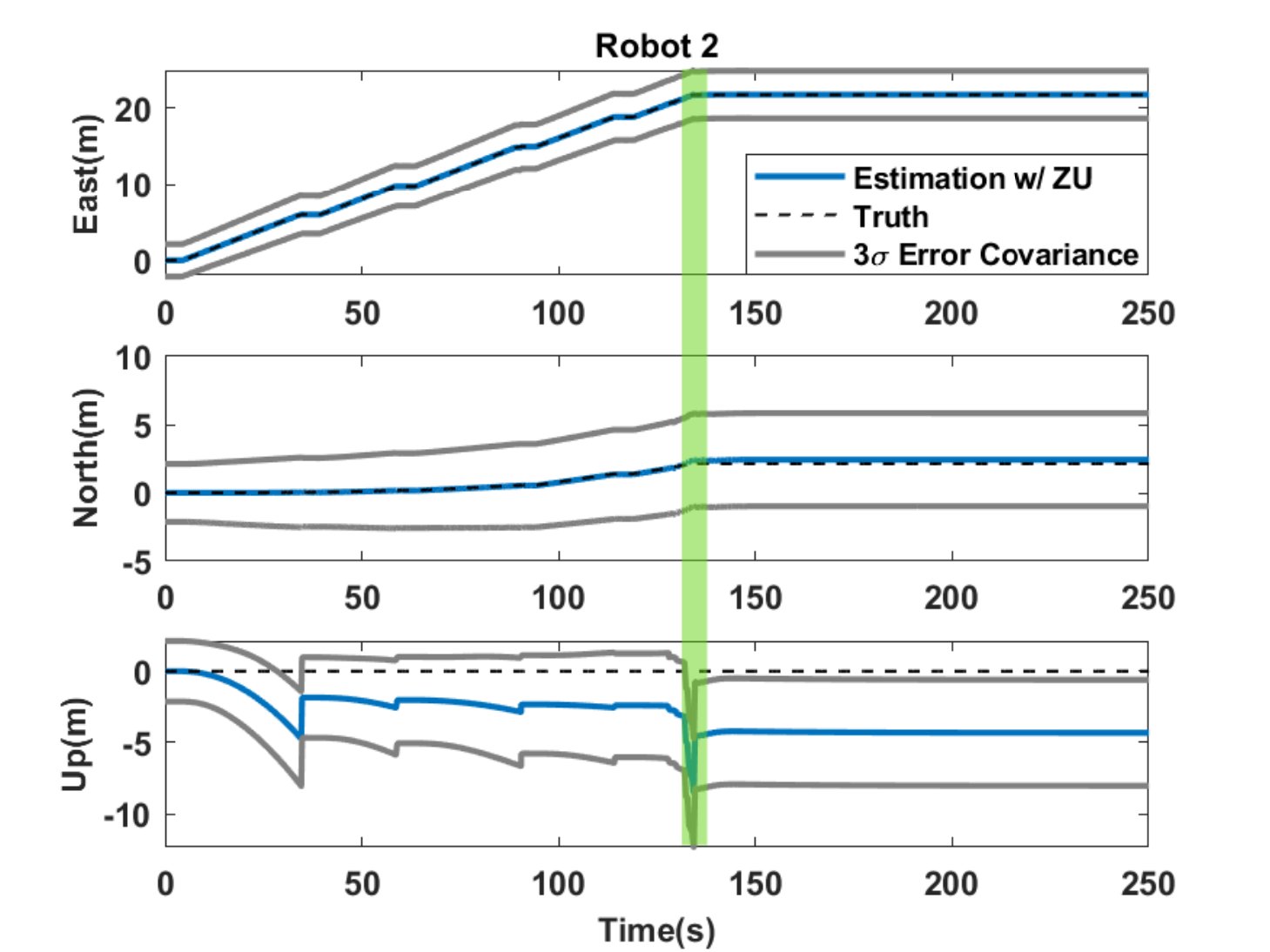}
		\label{fig:robot2_sim}
	}
	{
		\includegraphics[width=0.33\textwidth]{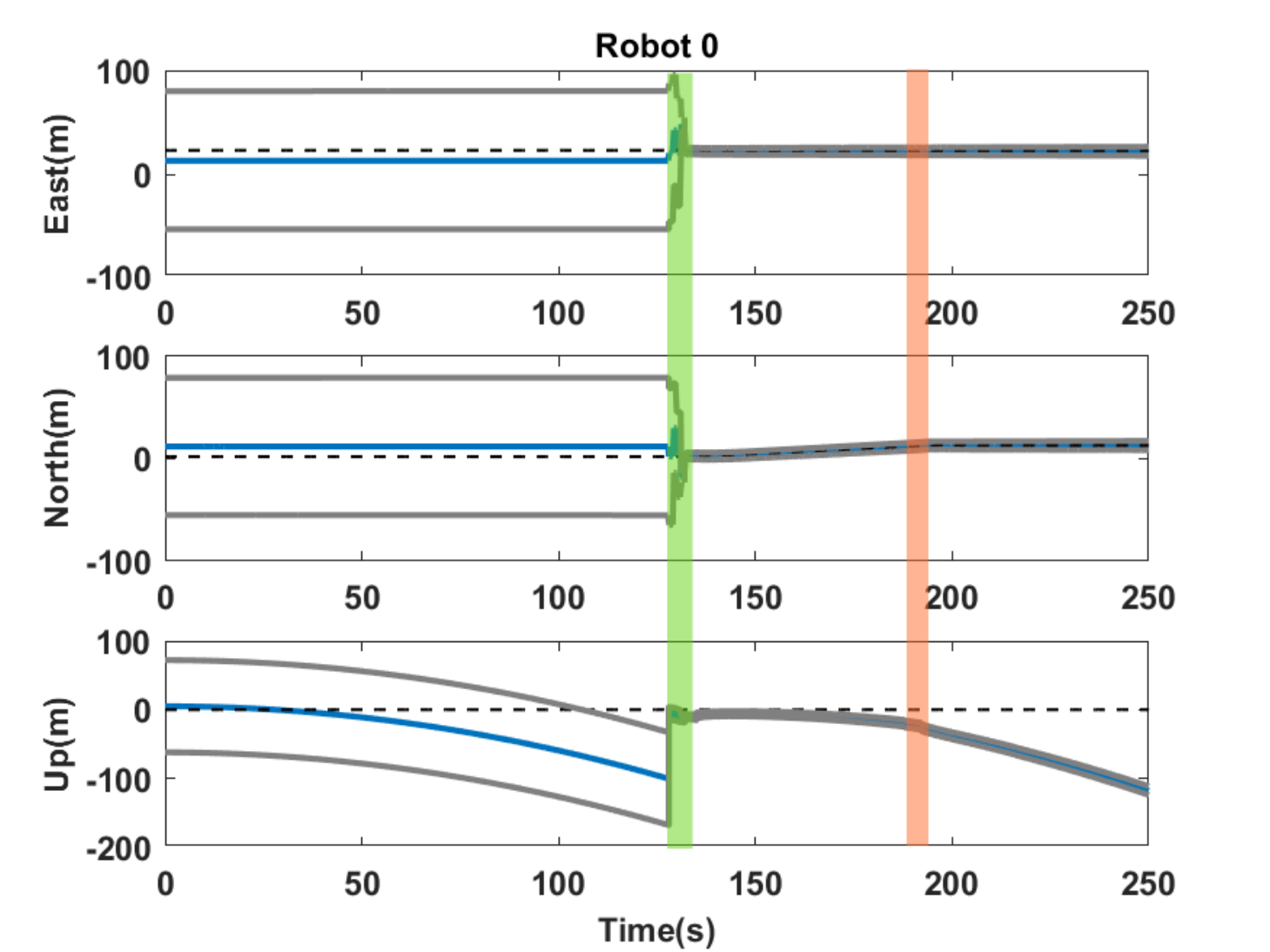}
		\label{fig:robot0_sim}
	}
	{
		\includegraphics[width=0.33\textwidth]{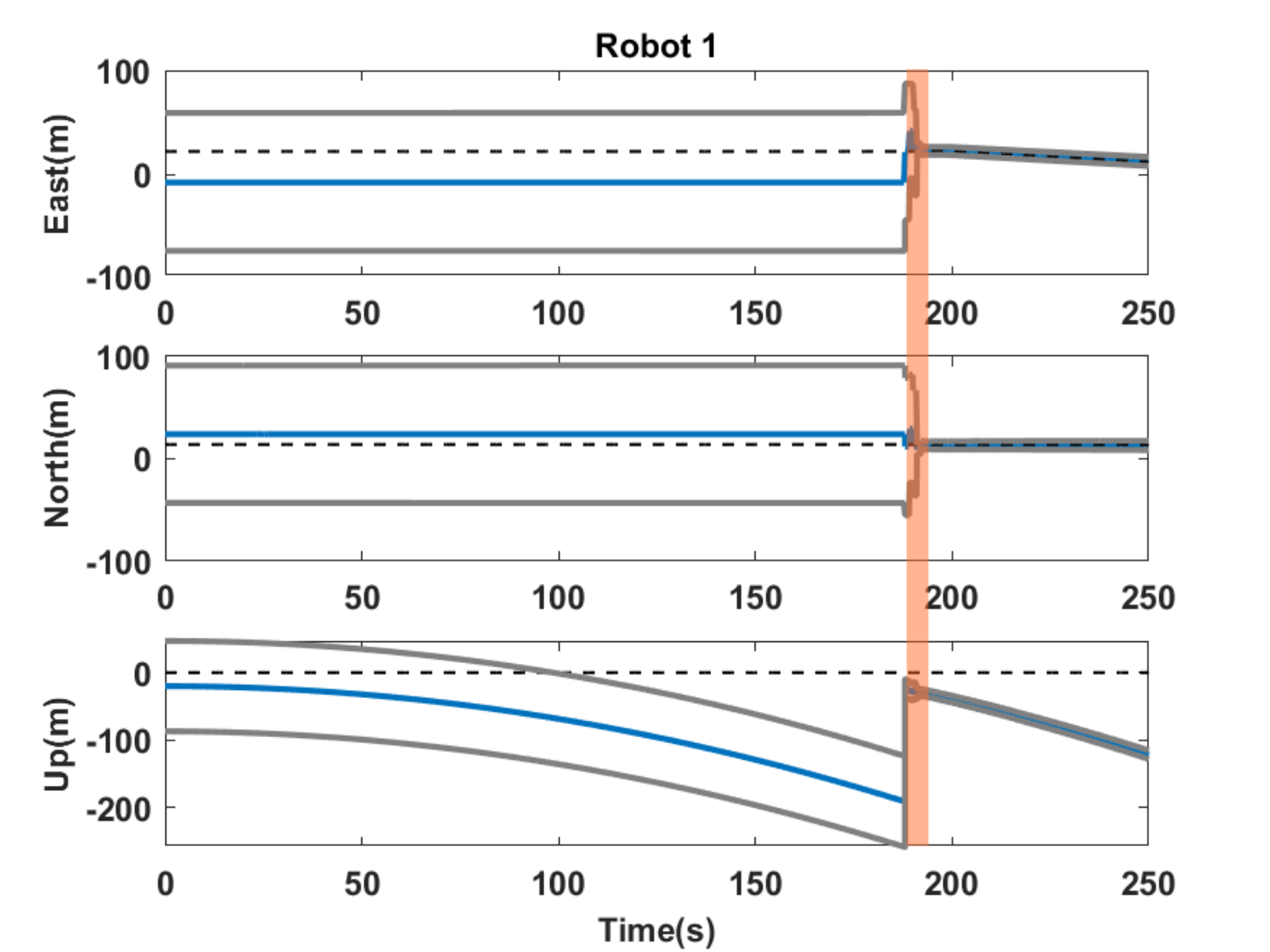}
		\label{fig:robot1_sim}
	}
\caption{East, North, and Up position estimation performance comparison for the cases where ZU is used versus not used by Robot 2 in simulation experiment (Test 1). The red dots represent the position estimation when ZU is not used, the blue dots represent the position estimation when ZU is used. The green shaded areas show the approximate duration when Robot 2 performs a relative update with Robot 0, and the orange shaded area show when Robot 0 performs a relative update with Robot 1. }    
\label{fig:simulation_2}
\end{figure}

To better visualize the effects of ZU, the position estimations of the robots in ENU frame with the 3$\sigma$ error covariance values and the truth are given in Fig.~\ref{fig:simulation_2}. It can be observed that the error for Robot 2 does not significantly change even when performing relative updates with lost robots. This is because of the filter properly considers which robot's localization to believe more based on the scale of the covariance. Leveraging ZU increases the effectiveness of relative update between robots. For example, since Robot 0 starts the test with a wrong estimation and high covariance, the position error and the covariance can only be reduced when this robot is able to perform a relative update with Robot 2. This also affects the Robot 1's localization performance. The dominant errors in the vertical direction can be primarily attributed to the fact that the robots are able to estimate their positioning based on IMU and WO measurements. While WO measurements provide information about the robot's translational motion on the x and y axes, they do not account for the vertical motion (z-axis) directly. The robots rely on IMU measurements to estimate their positioning in the vertical direction. As a result, the localization system is inherently more dependent on IMU measurements for the estimation of the vertical position, making it more susceptible to errors due to sensor noise, biases, and inaccuracies in the gravity model. Other factors that can contribute to the vertical estimation error include the alignment of the gravity vector with the vertical axis, which can lead to increased drift in vertical measurements. This drift may be more pronounced in the vertical direction compared to horizontal ones due to the inherent sensitivity of vertical motion to inaccuracies in the gravity model and sensor biases. Imperfect gravity modeling can also contribute to the observed vertical error. Using a more accurate gravity model and a simulated IMU sensor model could help reduce the vertical error in the estimation. 

\subsection {Real-World Experiment Results}

To further verify and evaluate the algorithm, we analyze the localization performance of Robot 2 in an indoor environment with real robots and real sensors for the cases where the ZU is being leveraged through its traversal or not. In Table~\ref{tab:comparison2}, the 3D position error for both cases are shown. In the same table, the average improvement is also shown for each metric. Applying ZU can bound the velocity error, calibrate IMU sensor biases, and limit the rate of INS localization error growth, even in a leveled, low-slip environment and short-distance scenarios such as ours, the improvement is a clear indication of the effectiveness of keeping localization estimation reliable by using ZU for a single robot. This can also be observed in Table \ref{tab:comparison2}.

\begin{table}[htb]
\centering
\footnotesize
\caption{Localization estimation comparison of Robot 2 in real world experiments }
\label{tab:comparison2}

\begin{tabular}{@{}clccc@{}}
\hline
& \textbf{Robot 2}&  \multicolumn{1}{c}{w/o ZU} & \multicolumn{1}{c}{w/ ZU} & \multicolumn{1}{c}{Improvement*}\\
\hline
\multirow{6}{*}{\rotatebox[origin=c]{90}{{\centering Error (m)}}}&
$\text{Max}_{E}$      & 2.22 & 1.60 & 27.76 \% \\
&$\text{Max}_{N}$         & 2.24 & 1.86 & 16.98 \% \\
&$\text{Max}_{U}$          & 58.21 & 2.11 & 96.38 \% \\
&$\text{RMSE}_{E} $       & 1.26 & 0.94 & 25.23 \% \\
&$\text{RMSE}_{N}$        & 1.26 & 1.03 & 18.19 \% \\
&$\text{RMSE}_{U}$        & 23.57 & 0.95 & 95.96 \% \\
\hline
\end{tabular}

\vspace{0.1cm}

\begin{tabular}{@{}clccc|ccc|cc@{}}
\hline
& & \multicolumn{3}{c}{w/o ZU} & \multicolumn{3}{c}{w/ ZU}\\ 
\hline
&\textbf{Robot 2 }& \scriptsize{Best}& \scriptsize{Worst}& \scriptsize{Average*}& \scriptsize{Best}& \scriptsize{Worst}& \scriptsize{Average*}& \scriptsize{Improvement*}\\
\hline
\multirow{4}{*}{\rotatebox{90}{\centering 3D Err (m)}}&
Median      & 6.27 & 20.79 & 13.06 & 0.61 & 2.39 & 1.57 & 87.98 \% \\
&Max         & 13.93 & 164.58 & 58.39 & 1.56 & 3.61 & 3.05 & 94.77 \% \\
&STD         & 4.37 & 47.37 & 15.83 & 0.34 & 0.91 & 0.85 & 94.66 \% \\
&RMSE        & 7.59 & 63.89 & 23.75 & 0.67 & 2.27 & 1.76 & 92.59 \% \\
\hline
\end{tabular}
\begin{tablenotes}
\centering
\item{*} The values are the average of 10 tests, improvement is based on the average values.
\end{tablenotes}
\end{table}

\begin{table} [h!]
\centering
\footnotesize
\begin{threeparttable}
\caption{Robot 1 and Robot 0 relative update performance analysis in real-world experiment }
\label{tab:r1r2relupd_rw}
\centering
\begin{tabular}{@{}clcc|cc|cc|cc@{}}
\hline
& & \multicolumn{4}{c}{Robot 1 Initial Error: 36.05m} & \multicolumn{4}{c}{Robot 0 Initial Error: 14.14}\\ 
\hline
& & \multicolumn{2}{c}{w/o ZU in Robot 2} & \multicolumn{2}{c}{w/ ZU Robot 2}& \multicolumn{2}{c}{w/o ZU in Robot 2} & \multicolumn{2}{c}{w/ ZU in Robot 2}\\ 
\hline
&\textbf{ }& \footnotesize{Correction (m)}& \footnotesize{Improvement (\%)}& \footnotesize{Correction (m)}& \footnotesize{Improvement (\%)}& \footnotesize{Correction (m)}& \footnotesize{Improvement (\%)}& \footnotesize{Correction (m)}& \footnotesize{Improvement (\%)}\\
\hline
\multirow{10}{*}{\rotatebox[origin=c]{90}{{\centering Horizontal Error }}}&
T1      & 33.29 & 92.32 \% & 35.85 & \textbf{99.42} \% & 11.12 & 78.64 \% & 13.70 & \textbf{96.88} \% \\
&T2     & 33.23 & \textbf{92.16} \% & 32.79 & 90.94 \% & 11.73 & \textbf{82.94} \% & 10.66 & 75.40 \% \\
&T3     & 29.88 & 82.87 \% & 32.72 & \textbf{90.76} \% & 7.10 & 50.23 \% & 11.06 & \textbf{78.20} \% \\
&T4     & 21.90 & 60.73 \% & 31.70 & \textbf{87.91} \% & 10.33 & \textbf{73.08} \% & 10.33 & 73.04 \%  \\
&T5     & 30.05 & 83.33 \% & 33.35 &\textbf{ 92.49} \% & 7.80 & 55.18 \% & 9.97 & \textbf{70.49} \%\\
&T6     & 32.11 & 89.07 \% & 33.16 & \textbf{91.96} \% & 9.14 & 64.67 \% & 10.36 & \textbf{73.24} \%  \\
&T7     & 33.55 &\textbf{ 93.05} \% & 32.97 & 91.43 \% & 11.56 & \textbf{81.76} \% & 10.28 & 72.69 \% \\
&T8     & 32.66 & 90.58 \% & 34.14 & \textbf{94.70} \% & 9.88 & 69.87 \% & 11.42 &\textbf{ 80.75} \% \\
&T9     & 30.46 & 84.49 \% & 33.67 & \textbf{93.38} \% & 6.93 & 49.02 \% & 10.79 & \textbf{76.33} \% \\
&T10    & 31.69 & 87.88 \% & 34.12 & \textbf{94.63} \% & 7.90 & 55.84 \% & 12.97 & \textbf{91.69} \%\\
\hline
\end{tabular}
\end{threeparttable}
\end{table}
\begin{figure}[h!] 
{
	{
		\includegraphics[width=0.33\textwidth]{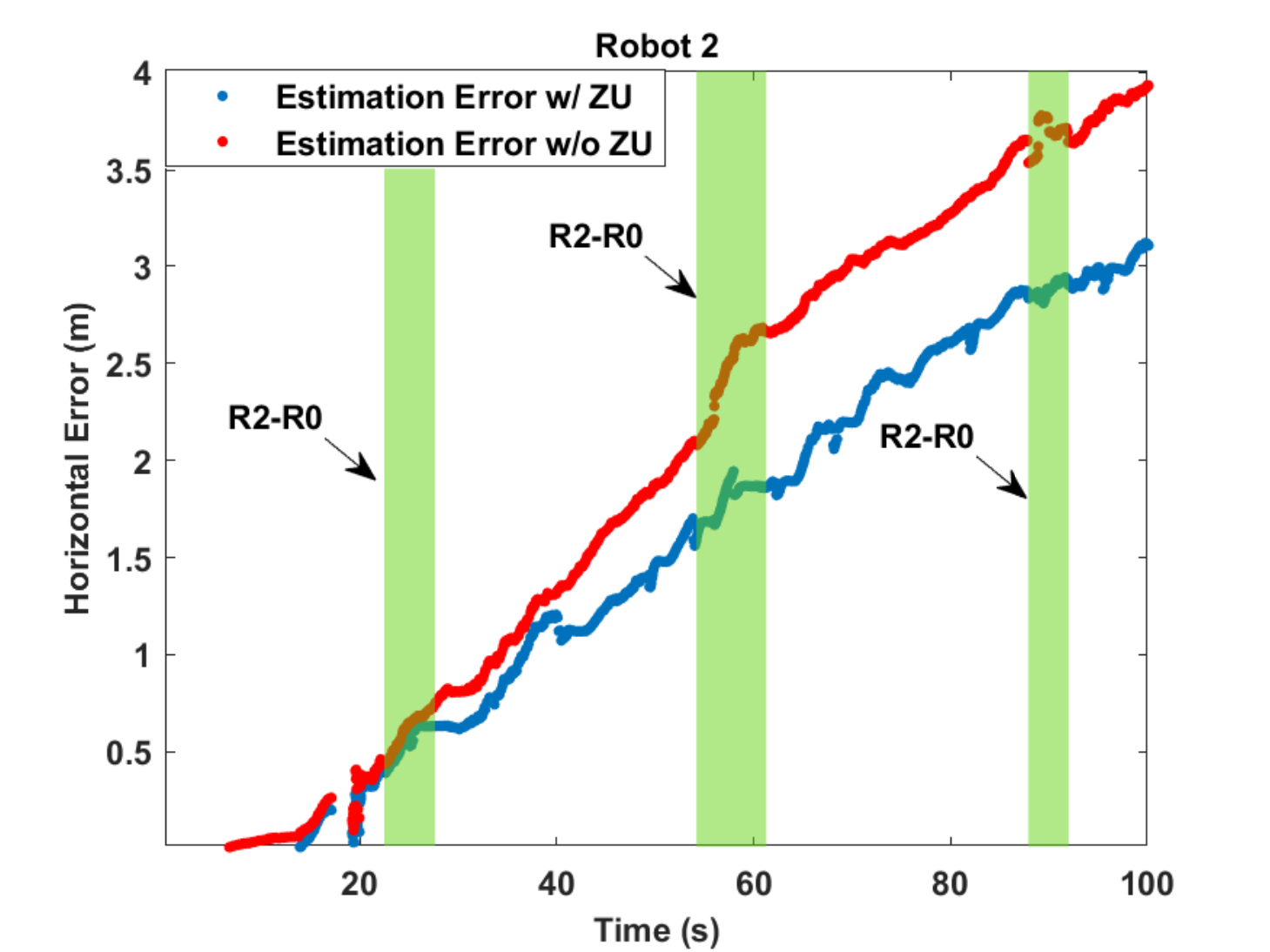}
		\label{fig:robot2}
	}
	{
		\includegraphics[width=0.33\textwidth]{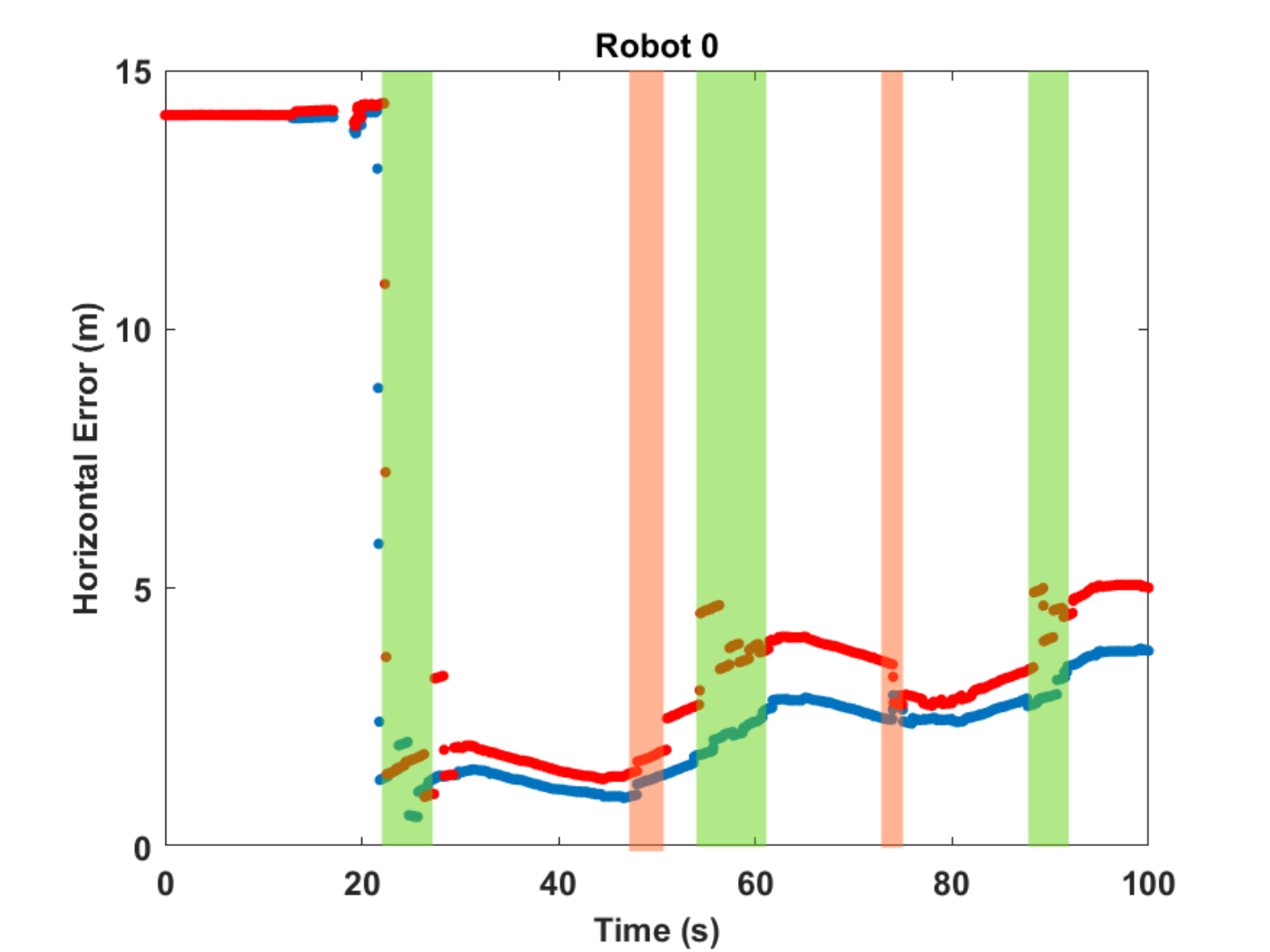}
		\label{fig:robot0}
	}
	{
		\includegraphics[width=0.33\textwidth]{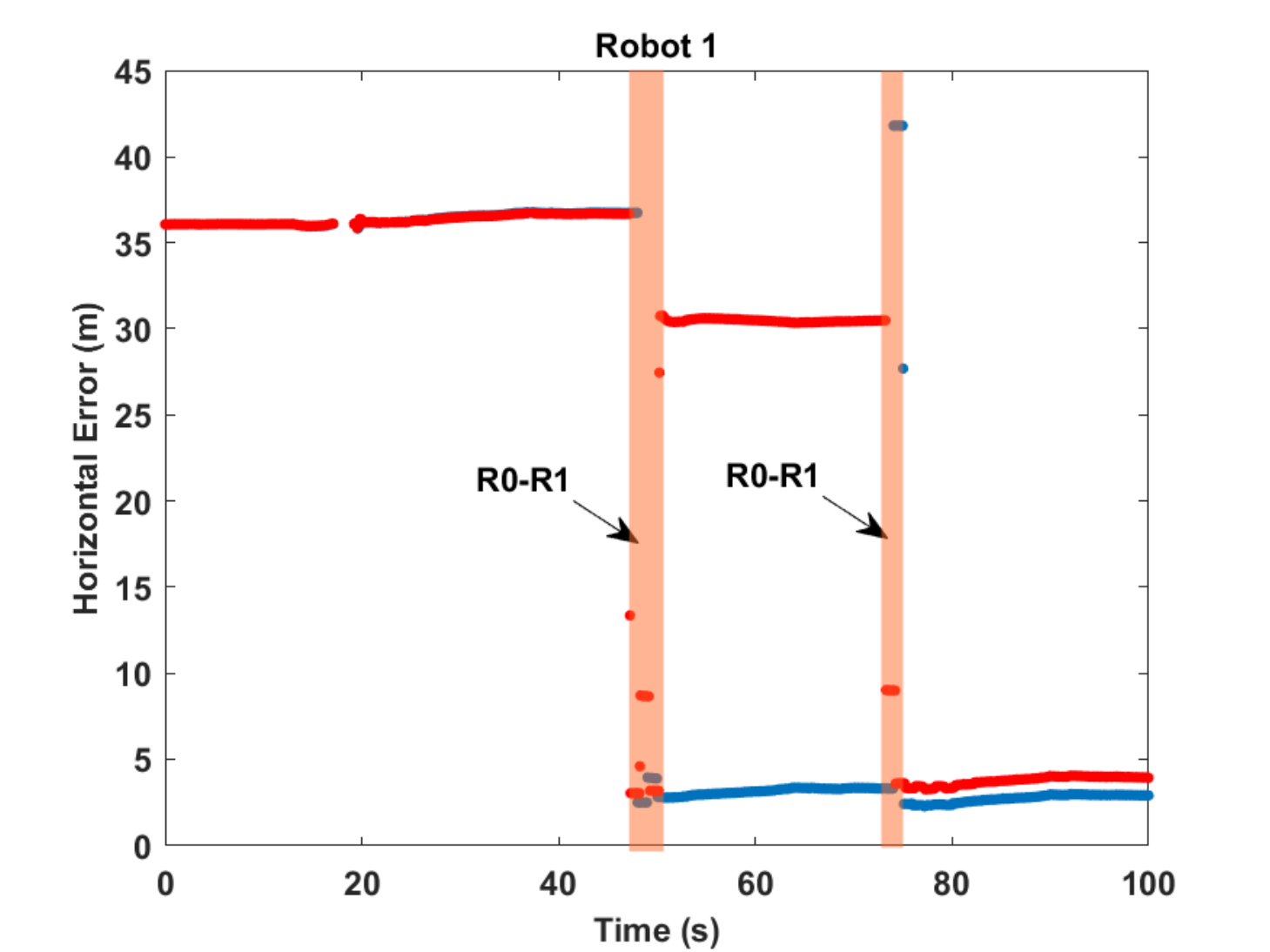}
		\label{fig:robot1}
	}
}    
\caption{Horizontal error estimation performance comparison for the cases where ZU is used versus not used by Robot 2 in real world experiment (Test 6). The red dots represent the estimation error when ZU is not used, the blue dots represent the estimation error when ZU is used. The green shaded areas show the approximate duration when Robot 2 performs relative update with Robot 0, and the orange shaded area show when Robot 0 performs relative update with Robot 1. }    
\label{fig:horzerrors_rw}
\end{figure}

As in the simulation results, the real-world experiments show a similar trend on correcting the errors, such that the benefit of using ZU by one robot is improving the overall localization performance of the multi-robot system. In Fig.~\ref{fig:horzerrors_rw}, the same color coded representation is used as in simulation experiments (i.e., red dots represent the estimation error without using ZU while the blue dots represent the estimation error using ZU in Robot 2). The experiment starts with a good initial estimation and smaller covariance error. It can be observed in Fig.~\ref{fig:horzerrors_rw} that the horizontal error for Robot 2 stays stable even when performing relative update with lost robots. This is because the covariance of the lost robots is much higher than the covariance of the Robot 2. The benefit of ZU can be clearly observed, since robots are able to exchange better information whenever one of them is able to perform ZU. 

\begin{figure}[htb] 

	{
		\includegraphics[width=0.33\textwidth]{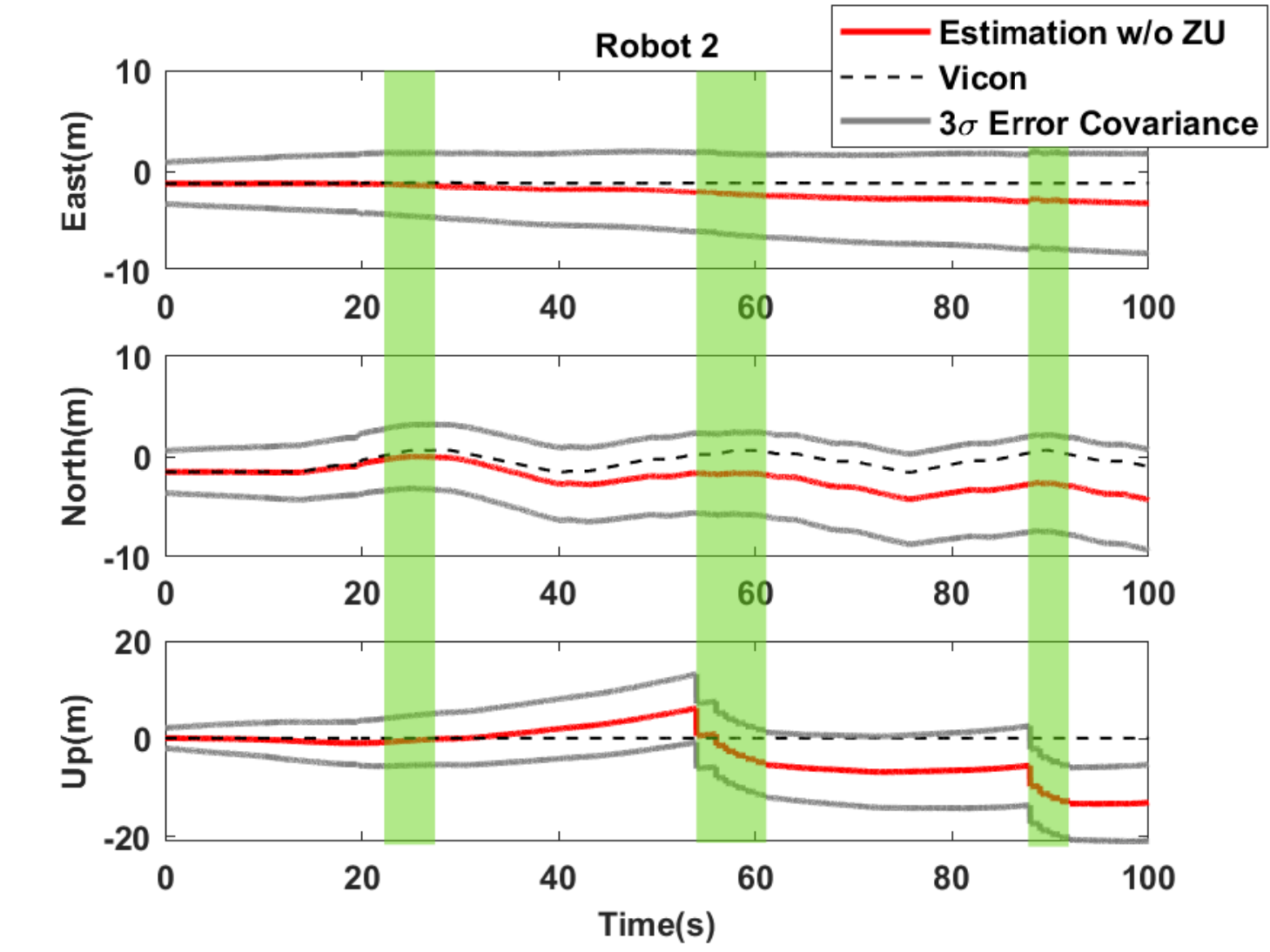}
		\label{fig:robot2}
	}
	{
		\includegraphics[width=0.33\textwidth]{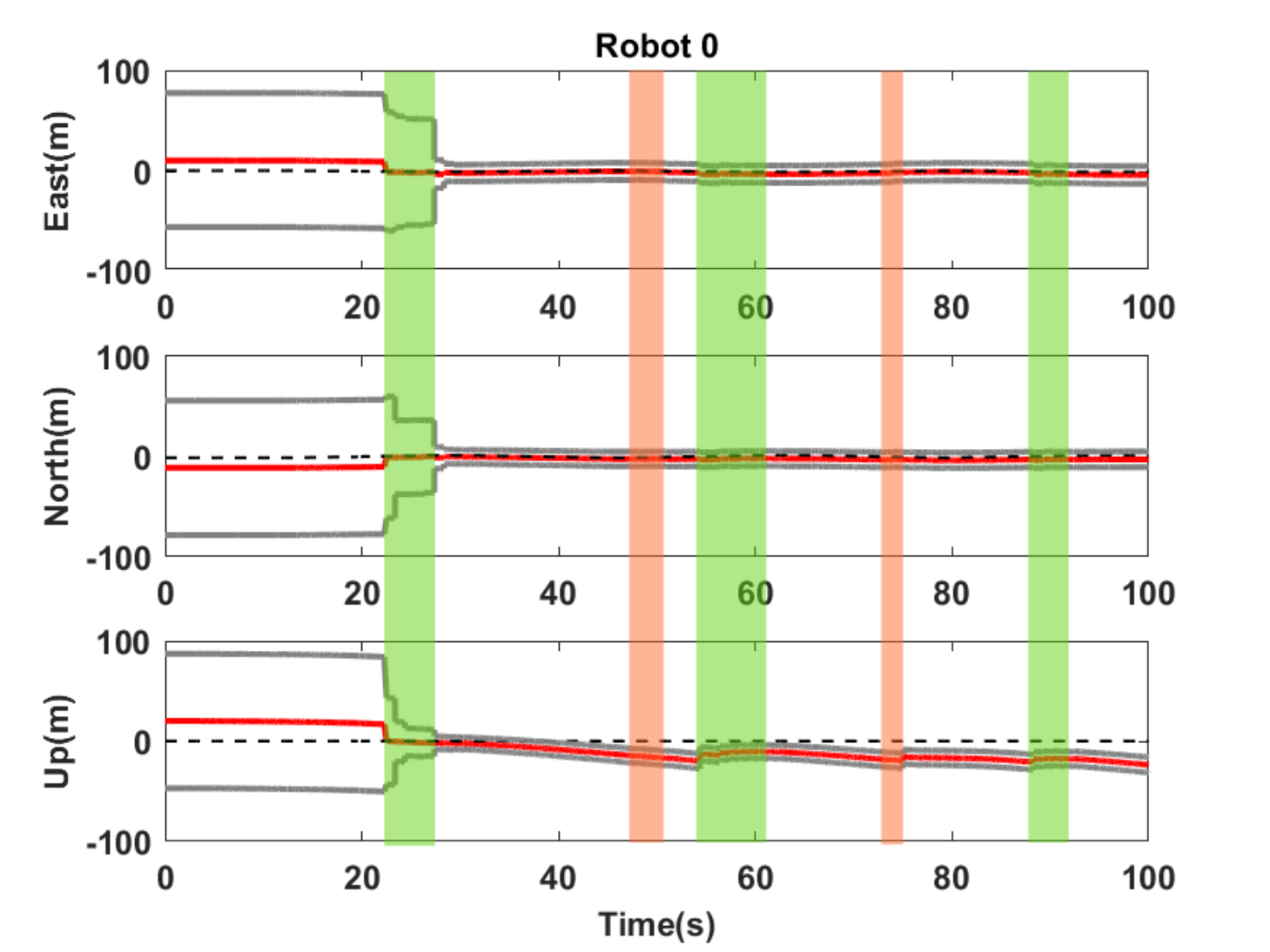}
		\label{fig:robot0}
	}
	{
		\includegraphics[width=0.33\textwidth]{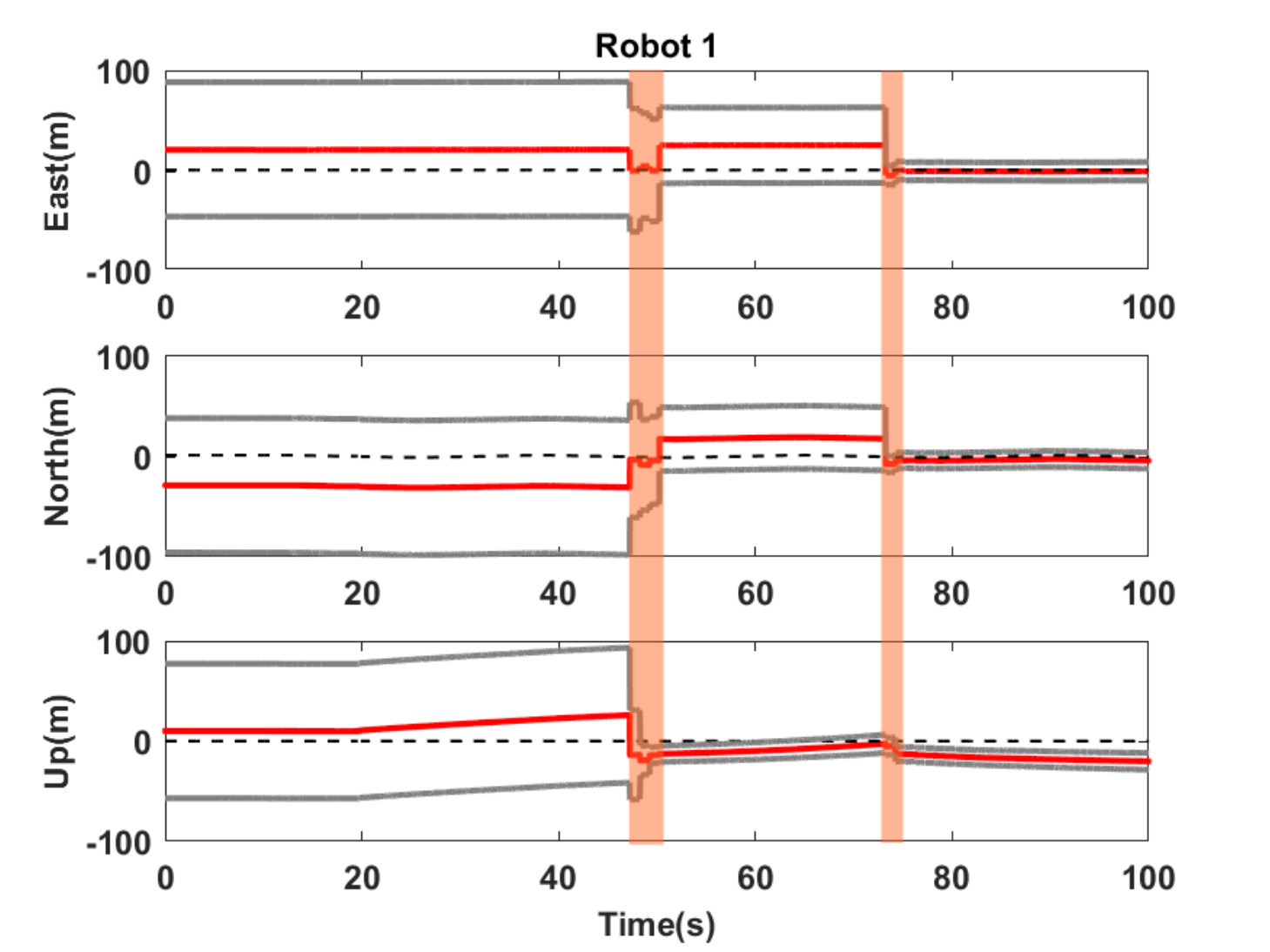}
		\label{fig:robot1}
	}
	
	{
		\includegraphics[width=0.33\textwidth]{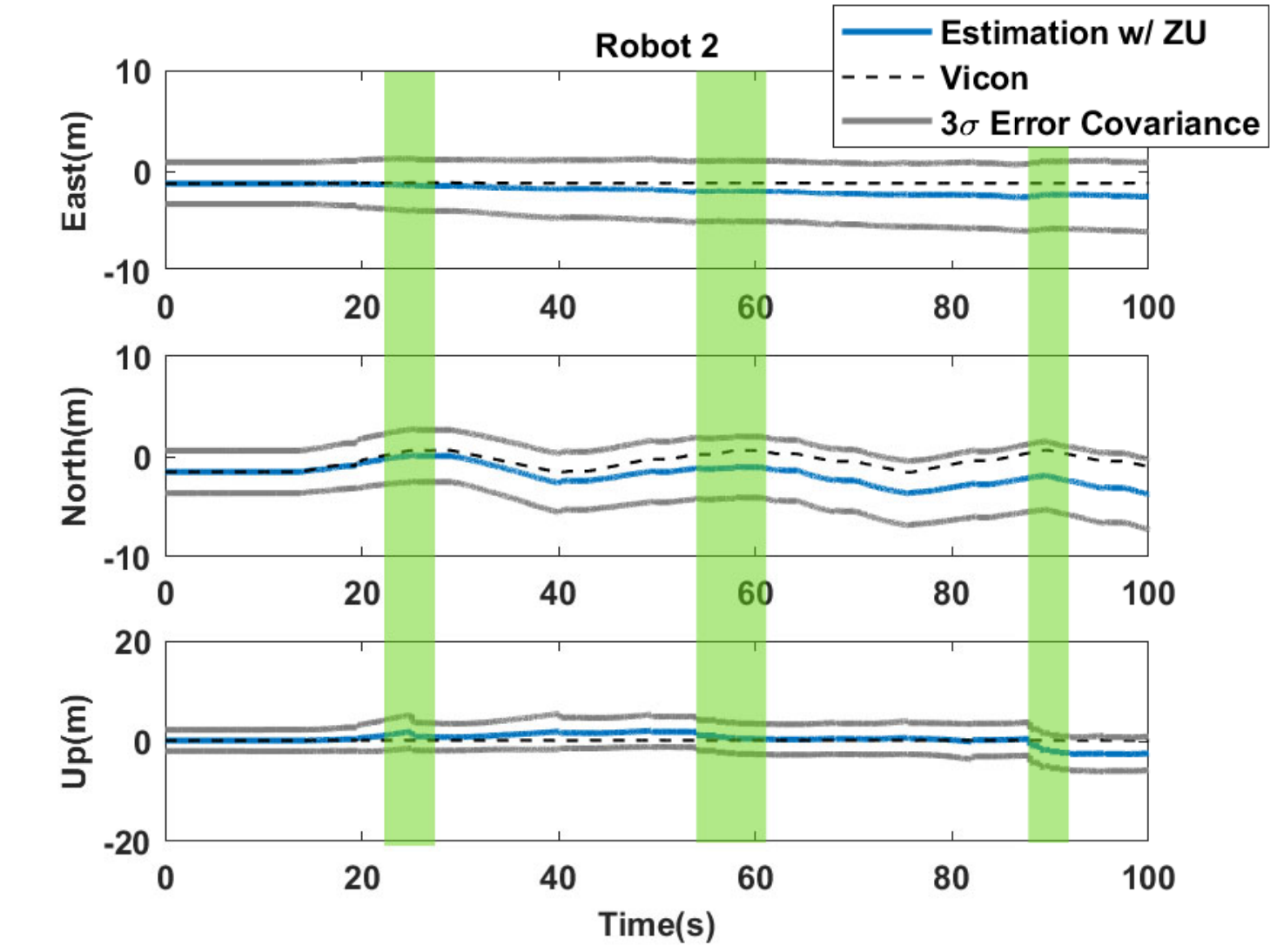}
		\label{fig:robot2}
	}
	{
		\includegraphics[width=0.33\textwidth]{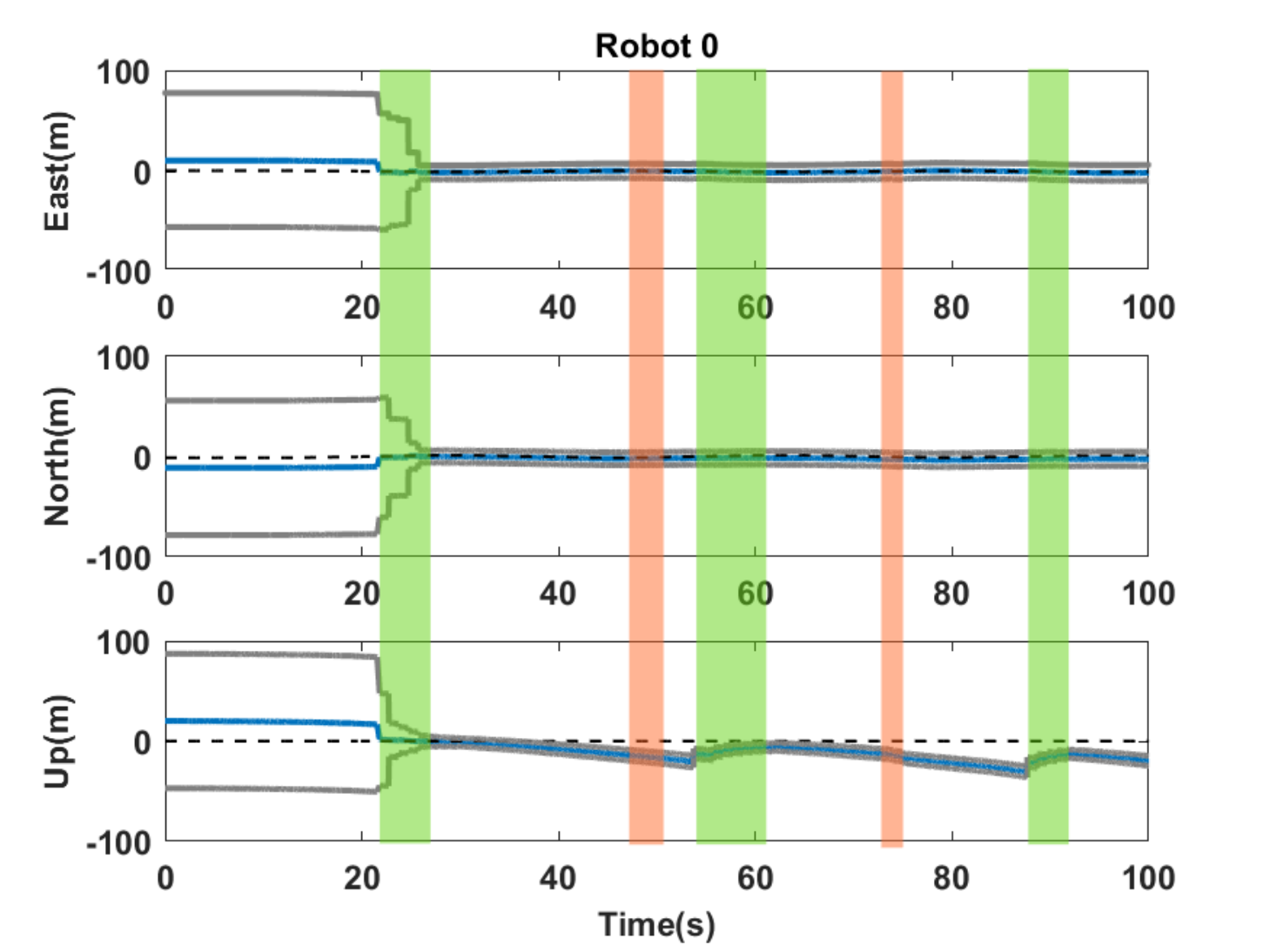}
		\label{fig:robot0}
	}
	{
		\includegraphics[width=0.33\textwidth]{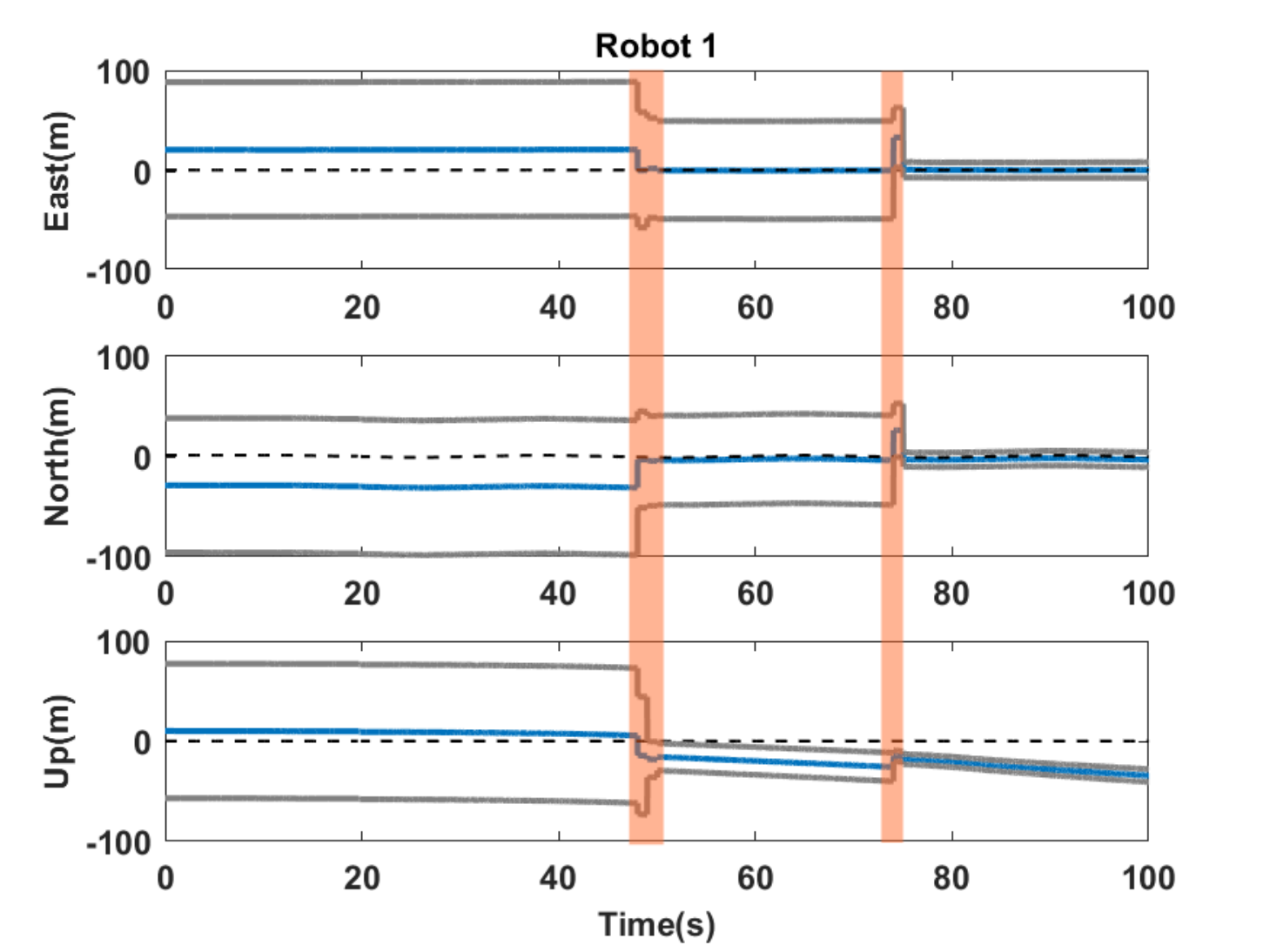}
		\label{fig:robot1}
	}
\caption{East, North, and Up (ENU) position estimation performance comparison for the cases where ZU is used versus not used by Robot 2 in real world experiment (Test 6). The red dots represent the position estimation when ZU is not used, the blue dots represent the position estimation when ZU is used. The green shaded areas show the approximate duration when Robot 2 performs a relative update with Robot 0, and the orange shaded area show when Robot 0 performs a relative update with Robot 1.  }    
\label{fig:realworld_enu}
\end{figure}

The position estimations of the robots in ENU frame with the 3$\sigma$ error covariance values and the truth are given in Fig.~\ref{fig:realworld_enu} to visualize the covariance changes during relative updates. For example, since Robot 0 starts the test with a wrong estimation and high covariance, the position error and the covariance can only be reduced when this robot is able to perform a relative update with Robot 2. This also affects the Robot 1's localization performance. It can also be seen that the position error in the $U$-axis is largely reduced when Robot 2 uses ZU, which leads to a significant amount of correction after relative update with Robot 0 in the same axis. Since the used robots in this study can only constrain the velocity in $E$ and $N$ axes by using wheel encoder measurements, the position error in $U$-axis can only be reduced by either ZU or relative update.

\section{Conclusion}
\label{Conclusion}

In this paper, we have proposed an error-state DEKF algorithm for cooperative localization of mobile robots in GNSS-denied/degraded environments using ZU, IMU, UWB, and odometry measurements. This work significantly expands upon our previous work in~\citet{edu2022pseudo}, generalizing the utilization of ZU to improve the localization performance in a DEKF based cooperative localization system. The proposed algorithm was implemented and tested with real hardware in a video motion capture facility and a ROS-based simulation environment for unmanned ground vehicles (UGV), aiming to re-localize the lost robots in the system.
The main contributions of this work are: (1) a novel method to leverage ZU in a decentralized cooperative localization framework, (2) the integration of odometry velocity measurements into the DEKF algorithm, (3) the use of ZU for reinstating lost robots in a multi-robot system, and (4) the real-world validation of the algorithm with multiple robots. Analyses and results demonstrate that using ZU in a cooperative D-EKF algorithm greatly benefits the localization estimation performance, making it a potential failsafe condition for other methods that might fail or be unreliable, such as in warehouse stocking, factory automation, and retail spaces. 

While ZU provides significant benefits to localization, it is worth noting potential misuse scenarios. For instance, misuse could occur if ZU is overly relied upon in environments where determining stationary conditions may be challenging. Overuse of ZU could also potentially affect the robot's traversability rate. Additionally, the necessity and impact of ZU depend on the quality and availability of external aids. In high-quality systems with uninterrupted positioning data (e.g., high-end GNSS, lidar SLAM), the reliance on ZU may decrease. However, in environments with unreliable or absent external aids, using ZU becomes more effective, as highlighted in this study featuring a system equipped with an IMU, wheel encoders, and UWB.

For future work, we plan to apply different constraints (e.g., non-holonomicity, hovering, landing) for other locomotion types which will allow for observing the performance of the algorithm in various situations. Additionally, we plan to incorporate obstacle avoidance strategies to ensure the safety of the robots and prevent any potential collisions, especially in cases where the position uncertainty is larger than the UWB detection range. Moreover, exploring the use of other types of sensors, such as cameras or lidars, to further enhance localization performance, and applying the proposed method to different types of robots, such as aerial or underwater vehicles, could be investigated. Finally, incorporating adaptive stopping strategies (e.g., determining optimal frequency and duration of stopping) for robots to perform ZU and using machine learning techniques to optimize the cooperative localization performance are potential avenues for further research. 

\section*{Acknowledgments}
This work was supported in part through a subcontract with Kinnami Software Corporation under the STTR project FA864921P1634. The authors thank to Dr. Yu Gu for allowing us to use the instrumented iRobot Create platforms; Jonas Bredu and Shounak Das for assisting with the tests.



\section*{Conflict of interest}

The authors declare no potential conflict of interests.



\printbibliography

\end{document}